\documentclass[preprint,11pt]{elsarticle}

\providecommand{\pgfsyspdfmark}[3]{}
\usepackage[utf8]{inputenc}
\usepackage[T1]{fontenc}
\usepackage[english]{babel}
\usepackage{amsmath,amsthm} 
\usepackage{mathtools}
\usepackage{dsfont}
\usepackage{empheq}
\usepackage{wasysym}
\usepackage{enumitem}
\usepackage{algorithm}
\usepackage{algorithmic}
\usepackage{natbib}
\usepackage{url}
\usepackage{ulem}
\usepackage[super]{nth}

\usepackage[toc,page]{appendix}
\usepackage[usenames,dvipsnames]{xcolor}
\usepackage[citecolor=Emerald,linkcolor=   NavyBlue,urlcolor=Magenta,colorlinks=true]{hyperref}

\makeatletter

\newskip\@bigflushglue \@bigflushglue = -100pt plus 1fil

\def\bigcentering{\let\\\@centercr\rightskip\@bigflushglue%
	\leftskip\@bigflushglue
	\parindent\z@\parfillskip\z@skip}

\makeatother

\usepackage{graphicx}

\usepackage{amssymb}

\theoremstyle{example}
\newtheorem{example}{Example}[section]

\theoremstyle{remark}

\theoremstyle{definition}
\newtheorem{definition}{Definition}[section]

\journal{ }

\begin{document}
	
	\begin{frontmatter}

		\title{Imputation and low-rank estimation with Missing Not At Random data}
		
		\author[label1,label2]{Aude Sportisse}
		\author[label1,label3]{Claire Boyer}
		\author[label2,label4]{Julie Josse}
		\address[label1]{Laboratoire de Probabilités Statistique et Modélisation, Sorbonne Universit\'e, France}
		\address[label3]{Département de Mathématiques et applications, Ecole Normale Supérieure, Paris, France}
		\address[label2]{Centre de Mathématiques Appliquées, Ecole Polytechnique, France}
		\address[label4]{XPOP, INRIA, France}
		
		\begin{abstract}
			Missing values challenge data analysis because many supervised and unsupervised learning methods cannot be applied directly to incomplete data.
			Matrix completion based on low-rank assumptions are very powerful solution for dealing with missing values. However, existing methods do not consider the case of informative missing values which are widely encountered in practice. 
			This paper proposes matrix completion methods to recover Missing Not At Random (MNAR) data. 
			Our first contribution is to suggest a model-based estimation strategy by modelling the missing mechanism distribution. An EM algorithm is then implemented, involving a Fast Iterative Soft-Thresholding Algorithm (FISTA).
			Our second contribution is to suggest a computationally efficient surrogate estimation by implicitly taking into account the joint distribution of the data and the missing mechanism:  the data matrix is concatenated with the mask coding for the missing values; a low-rank structure for exponential family is assumed on this new matrix, in order to encode links between variables and missing mechanisms.  The methodology that has the great advantage of handling different missing value mechanisms is robust to model specification errors. 
			
			The performances of our methods are assessed on the real data collected from a trauma registry (TraumaBase$^{\mbox{\normalsize{\textregistered}}}$) containing clinical information about over twenty thousand severely traumatized patients in France. The aim is then to
			predict if the doctors should administrate tranexomic
			acid to patients with traumatic brain injury, that would limit excessive bleeding.
		\end{abstract}
		
		
	\end{frontmatter}
	
	
	\section{Introduction}
	\label{Section1}

The problem of missing data is ubiquitous in the practice of data analysis. Main approaches for handling missing data include imputation methods  and  the use of Expectation-Maximization (EM) algorithm \citep{dempster1977maximum} which allows to get the maximum likelihood estimators in various incomplete-data problems \citep{rubin_little}. The theoretical guarantees of these methods ensuring the correct prediction of missing values or the correct estimation of some parameters of interest are only valid if some assumptions are made on how the data came to be missing. \citet{rubin1} introduced three types of missing-data mechanisms:  (i) the restrictive assumptions of missing completely at random (MCAR) data, (ii) the missing at random (MAR) data, where the missing data may only depend on the observable variables, and  (iii) the more general assumption of missing not at random (MNAR) data, \textit{i.e.} when the unavailability of the data depends on the values of other variables and its own value. A classic example of MNAR data, which is the focus of the paper, is surveys where rich people would be less willing to disclose their income or where people would be less incline to answer sensitive questions on their addictive use. 
Another example would be the diagnosis of Alzheimer's disease, which can be made using a score obtained by the patient on a specific test. However, when a patient has the disease, he or she has difficulty answering questions and is more likely to abandon the test before it ends. 

\paragraph{Missing non at random data}
When data are MCAR or MAR, valid inferences can be obtained by ignoring the missing-data mechanism \citep{rubin_little}. The MNAR data lead to selection bias, as the observed data are not representative of the population. In this setting, the missing-data mechanism must be taken into account, by considering the joint distribution of complete data matrix and the missing-data pattern. There are mainly two approaches to model the joint distribution using different factorizations: 
\begin{enumerate}
	\item selection models \citep{heckman1974sample},  which seem preferred as it models the distribution of the data, say $Y$, and the incidence of missing data as a function of $Y$ which is rather intuitive;
	\item pattern-mixture models \citep{little1993pattern}, which key issue is that it requires to spe\-ci\-fy the distribution of each missing-data pattern separately. 
\end{enumerate}
Most of the time, in these parametric approaches, the EM algorithm  is performed to estimate the parameters of interest, such as the parameters of generalized linear models in \cite{Ibrahim} and the missing-data mechanism distribution is usually specified by  logistic regression models \citep{Ibrahim, tang2017random,morikawa2017semiparametric}, in the case of selection models. In addition, the MNAR mechanism often is chosen self-masked \textit{i.e.} the lack of a variable depends only on the variable itself and only simple models have been considered with cases where just the output variable or one or two variables are subject to missingness \citep{miao2017identification,Ibrahim}. Note that recent works based on graph-based approaches \citep{Mohan2018graphical, mohan2018estimation} show that in some specific setting of MNAR values, it is possible to estimate parameters for simple models, such as the mean and variance in linear models, without specifying the missing value mechanism.

\paragraph{Low-rank models with missing values}
In this paper, we focus on estimation and imputation in low-rank 
models with MNAR data. The low-rank model has become very popular in recent years \citep{kishore2017literature} and it plays a key role in many scientific and engineering tasks, including denoising \citep{gavish2017optimal}, collaborative filtering \citep{yang2018oboe}, genome-wide studies \citep{leek2007capturing, price2006principal}, and functional magnetic resonance imaging \citep{candes2013sure}. It is also a very powerful solution for dealing with missing values \citep{josse2016denoiser,kallus2018causal}.
Indeed, the low-rank assumption can be considered as an accurate approximation for many matrices as detailed in \cite{Udell2017logrank}. For instance, the low-rank approximation makes sense when either, one can consider that a limited number of individual profiles exist or, dependencies between variables can be established. 

Let us consider a data matrix $Y \in \mathbb{R}^{n\times p}$ which is a noisy realisation of a low-rank matrix $\Theta \in \mathbb{R}^{n\times p}$ with rank $r < \min \{n,p\}$:
\begin{equation}\label{model}
Y=\Theta+\epsilon, \textrm{where}
\left\{
\begin{array}{ll}
\Theta \textrm{ has a low rank $r$,} \\
\epsilon \sim \mathcal{N}(0,\sigma^2 I ).
\end{array}
\right.
\end{equation}
In the following, $\sigma$ is assumed to be known. Suppose that only partial observations are accessible. We note the mask $\Omega \in \{0,1\}^{n\times p}$ with 
$$\Omega_{ij}=\left\{
\begin{array}{ll}
0 \textrm{ if } y_{ij} \textrm{ is missing,} \\
1 \textrm{ otherwise.}
\end{array}
\right.$$
where $y$ is a realisation of $Y$. 
The main objective is then to estimate the parameter matrix $\Theta$ from the incomplete data, which can be seen on the one hand as a denoising task by estimating the parameters from the observed incomplete noisy data,  and on the other hand as a prediction task by imputing missing values with values given by the estimated parameter matrix.
A classical approach to estimate  $\Theta$ with  MAR or MCAR missing values  are based on convex relaxations of the rank, \textit{i.e.} the nuclear norm and consists in solving the following penalized  weighted least-squares problem:

\begin{equation}\label{OptimMAR}
\hat{\Theta} \in \textrm{argmin}_\Theta \|(Y-\Theta) \odot \Omega \|_F^2 + \lambda \|\Theta\|_\star,
\end{equation}
where $\|.\|_F$ and $\|.\|_{\star}$ respectively denote the Frobenius norm and the nuclear norm and $\odot$ is the Hadamard product.
The main algorithm available to solve \eqref{OptimMAR} consists in a proximal gradient method, leading to iterative soft-thresholding algorithm (ISTA) of the singular value decomposition (SVD) \citep{mazumder2010spectral, cai2010singular} in the case of a regularization via the nuclear norm (note that this strategy is equivalent to perform an EM algorithm with a nuclear norm penalization in the M-step, see Appendix \ref{sec:EMsoft}). Given any initialization (for instance the missing values can be initialized to the mean of the non-missing entries), a soft-thresholding SVD is computed on the completed matrix and the predicted values of the missing entries are updated using the values given by the new estimation. The two steps of estimation and imputation are iterated until empirical stabilization of the prediction. 
There has been a lot of work on denoising and matrix completion with low-rank models, whether algorithmic, methodological or theoretical contributions \citep{candes2009exact,candes2010matrix}. However, to the best of our knowledge most of the existing methods do not consider the case of MNAR data.

\paragraph{Contributions}
In order to perform low-rank estimation with MNAR data, our first contribution, detailed in Section \ref{sec:model}, is to suggest a model-based estimation strategy by maximizing the joint distribution of the data and the missing values mechanism using an EM algorithm. More specifically, a Monte Carlo approximation is performed coupled with the Sampling Importance Resampling (SIR) algorithm.  Note yet that introducing such a model for MNAR data does not prevent from handling Missing Completely At Random (MCAR) or Missing At Random (MAR) data as well. Indeed, our model can only impact variables of type MNAR, while the low-rank assumption will be enough to deal with other types of missing variables. This approach, although theoretically sound and well defined, has two drawbacks: its computational time and the need to specify an explicit model for the mechanism, so to have a strong prior knowledge about the shape of the missing-data distribution.  
\\
Our second contribution (Section \ref{sec:mask})  is to suggest an efficient surrogate estimation by implicitly modelling the joint distribution. To do so, we suggest to concatenate the data matrix and the missing-data mask, \textit{i.e.} the indicator matrix coding for the missing values, and to assume a low-rank structure on this new matrix in order to take into account the relationship between the variables and the mechanism. This strategy has the great advantage that it can be performed using classical methods used in the MCAR and MAR settings  and that it does not require to specify a model for the mechanism. This approach can be seen as connected to the following works. \cite{harel2009partial} presents a method to handle missing data in a latent-class model where the missing covariates $X$ are linked to the missing-data pattern $M$ by a latent variable $\eta$. In an example, they suggests treating $M$ as additional items alongside $X$, in order to make statistical inferences. Moreover, in the context of decision trees used for classification, \cite{twala2008good} suggests an approach known as missing values attribute where at each split, all the missing values can go on the right or on the left. This can be seen as cutting according to the missing value pattern so it is equivalent as implicitly adding $M$ with the covariates $X$. 
Finally, from the optimization point of view, we also suggest (Section \ref{sec:FISTA}) to use an accelerated proximal gradient algorithm, also called Fast Iterative
Soft-Thresholding Algorithm (FISTA) \citep{beck2009fast}  which is an accelerated version of the classical iterative SVD algorithm in the case of a penalization with the nuclear norm. 





\medbreak

The rest of the article is organized as follows. First, although the missing-data mechanism framework is widely used, there are points of ambiguity in the classical definitions, especially considering whether the statements hold for any value (from any sample) or for the realised value (from a specific sample) \citep{Seaman2013MAR, murray}. Therefore, Section \ref{Section1bis} is dedicated to specify a general and clear framework of the missing-data mechanisms in order to remove ambiguities and introduce the MNAR mechanism being considered. In Section \ref{Section2}, we present both proposals to address the MNAR data issue: by explicitly modelling the missing mechanism  or by implicitly taking it into account. Section \ref{Section3} is devoted to a simulation study on synthetic data. In Section \ref{sec:Traumabase}, we apply the model-based method to the TraumaBase$^{\mbox{\normalsize{\textregistered}}}$ dataset in order to
to assist doctors in making decisions about the administration of an active substance, called the tranexomic acid, to patients with traumatic brain injury. Finally, a discussion on the results and perspectives is proposed on Section \ref{sec:discussion}. 

\section{The missing-data mechanism: notations and definitions}
\label{Section1bis}

In the sequel,  we write the complete data matrix $Y\in \mathbb{R}^{n\times p}$ of quantitative variables,  whose distribution is parameterized by $\Theta$. 
The missing-data pattern is denoted by $M\in \{0,1\}^{n\times p}$ and $\phi$ is the parameter of the conditional distribution of $M$ given $Y$.
We assume the distinctness of the parameters, \textit{i.e.} the joint parameter space of $(\Theta,\phi)$ is the product of the parameter space of $\Theta$ and the one of $\phi$.  We start by writing the most popular definitions of \cite{rubin_little} for the missing-data mechanism.  By writing,  $Y=(Y_{\textrm{obs}},Y_{\textrm{mis}})$, where $Y_{\textrm{obs}}$ and $Y_{\textrm{mis}}$ denote the observed components and the missing ones of $Y$ respectively, they define:
\begin{align*}
p(M|Y;\phi)&=p(M; \phi), \quad \forall Y, \phi  &\textrm{(MCAR)} \\
p(M|Y;\phi)&=p(M|Y_{\textrm{obs}}; \phi), \quad  \forall Y_{\textrm{mis}}, \phi   &\textrm{(MAR)} \\
p(M|Y;\phi)&=p(M|Y_{\textrm{obs}},Y_{\textrm{mis}};\phi), \quad \forall \phi &\textrm{(MNAR)}
\end{align*}
Note that all matrices may be regarded as vectors of size $n \times p$ (see Example \ref{ex:Def}).
There are mainly two ambiguities: (i) it is unclear whether the equations hold for any realisation $(y,m)$ of $(Y,M)$, although it is widely understood as such and (ii)  $Y_\textrm{obs}$ and $Y_{\textrm{mis}}$ are actually functions of $M$, which is extremely confusing and explain why other attempts for definitions and notations are necessary. 
\cite{Seaman2013MAR} propose two definitions of the MAR mechanism, for which they differentiate if (i) the statements hold for any values (from any sample), the everywhere case (EC) (ii) or for the realised values (from a specific sample), the realised case (RC). They also introduce a specific notation for the observed values of $Y$, clearly written as a function $o$ of $Y$ and $M$: $o(Y,M)$. By writing  
$\tilde{y}$ and $\tilde{m}$ the realised values of $Y$ and $M$ for a specific sample, it leads to:
\begin{multline*}
\forall y,y^*,m \textrm{ such that } o(y,m)=o(y^*,m) \\
p(M=m|Y=y;\phi)=p(M=m|Y=y^*;\phi), \textrm{ (EC)}
\end{multline*}
\begin{multline*}
\forall y,y^* \textrm{ such that } o(y,\tilde{m})=o(y^*,\tilde{m})=o(\tilde{y},\tilde{m}) \\
p(M=\tilde{m}|Y=y;\phi)=p(M=\tilde{m}|Y=y^*;\phi), \textrm{ (RC)}
\end{multline*}
We can illustrate these concepts with the following 
example:
\begin{example}
	\label{ex:Def}
	Let $y=\left(\begin{matrix} 
	1 & 3 \\
	4 & 10
	\end{matrix}\right)
	$, that can be regarded as a vector $\mathrm{vec} (y)=\left(\begin{matrix} 
	1 & 3 & 4 & 10
	\end{matrix}\right)
	$. If $\mathrm{vec} (y)=\left(\begin{matrix} 
	1 & 3 & 4 & \textrm{NA}
	\end{matrix}\right)
	$ is observed, then $\tilde{m}=\left(\begin{matrix} 
	1 & 1 & 1 & 0 \end{matrix}\right)$ and $o(\tilde{y},\tilde{m})=\left(\begin{matrix} 
	1 & 3 & 4
	\end{matrix}\right)$. The data are realised MAR if 
	\begin{multline*}
	p(M=(1,1,1,0)|Y=y;\phi)\\
	=p(M=(1,1,1,0)|Y=y^*;\phi),\\
	\forall y,y^*, \: o(y,\tilde{m})=o(y^*,\tilde{m})=(1,3,4)
	\end{multline*}
	$$\Updownarrow$$
	\begin{multline*}
	p(M=(1,1,1,0)|Y=(1,3,4,a);\phi)\\
	=p(M=(1,1,1,0)|Y=(1,3,4,b);\phi),\forall a,b
	\end{multline*}
\end{example}
%


By extending the framework of \cite{Seaman2013MAR}, the MNAR mechanism can be defined in the everywhere case and with the two following assumptions: 
\begin{itemize}
	\item the missing-data indicators are independent given the data, 
	\item the MNAR mechanism is said to be self-masked, which assures that the distribution of a missing-data indicator $M_{ij}$ given the data $Y$ is a function of $Y_{ij}$ only. 
\end{itemize}

In the specific case of low-rank models, these both assumptions allow to have the independence by unit and to make the computations easier. 
\begin{definition}
	\label{def:selfevery}
	The missing data are generated by the self-masked everywhere MNAR mechanism if:
	$$p(M=\Omega|Y=y;\phi)=\prod_{i=1}^{n}\prod_{j=1}^{p}p(\Omega_{ij}|y_{ij};\phi),  \quad \forall Y, \phi$$
\end{definition}




\section{Proposition}
\label{Section2}

Our propositions for low-rank estimation with MNAR data require the following comments on the classical algorithms to solve \eqref{OptimMAR}. 
First, as in regression analysis there is an equivalence between minimizing least-squares and maximizing the likelihood under Gaussian noise assumption. 
Here as specified in Equation \eqref{model}, the entries $(Y_{ij})_{ij}$'s are assumed to be  independent and normally distributed, for all $ i \in \left[1,n\right],  j \in \left[1,p\right]$:
\begin{equation}
\label{eq:densitydata}
p(y_{ij};\Theta_{ij})=(2\pi\sigma^2)^{-1/2}e^{\left(-\frac{1}{2}\left(\frac{y_{ij}-\Theta_{ij}}{\sigma}\right)^2\right)}.
\end{equation}
It implies that we can show  (in Appendix \ref{sec:EMsoft}) that the classical proximal gradient methods to solve the penalized weighted least-squares criterion \eqref{OptimMAR}, such as iterative thresholding SVD, can be seen as a genuine EM algorithm, maximizing the observed penalized likelihood. 
Second, as detailed in Section \ref{sec:FISTA},  \eqref{OptimMAR} can be solved using a fast iterative soft-thresholding algorithm (FISTA) \citep{beck2009fast}.

\subsection{Modelling the mechanism} 
\label{sec:model} 
Considering the framework of selection models  \citep{heckman1974sample}, the first proposition consists in handling MNAR values in the low-rank model \eqref{model}, by specifying a distribution for the missing-data pattern $M$. Here, the missing data models $M_{ij}$ given the data $Y_{ij}$ are assumed to be independent and distributed by a logistic model, $\forall i \in \left[1,n\right], \forall j \in \left[1,p\right]$:
\begin{multline}
\label{densmecha}
p(\Omega_{ij}|y_{ij};\phi)=[(1+e^{-\phi_{1j}(y_{ij}-\phi_{2j})})^{-1}]^{(1-\Omega_{ij})} \\
[1-(1+e^{-\phi_{1j}(y_{ij}-\phi_{2j})})^{-1}]^{\Omega_{ij}}, 
\end{multline}
where $\phi_j=(\phi_{1j},\phi_{2j})$ denotes the parameter vector for conditional distribution of $M_{ij}$ given $Y_{ij}$ for all $i$.

Then, the joint distribution of the data and mechanism can be specified. Due to independence (see Definition \eqref{def:selfevery}): 
\begin{align*}
p(y,\Omega;\Theta,\phi)&=p(y;\Theta)p(\Omega|y;\phi) \\
&=\prod_{i=1}^{n}\prod_{j=1}^{p} p(y_{ij};\Theta_{ij})p(\Omega_{ij}|y_{ij};\phi_j).
\end{align*}
This leads to the joint negative log-likelihood: 
$$
\ell(\Theta,\phi;y,\Omega)=- \sum_{i=1}^{n} \sum_{j=1}^{p} \ell((\Theta_{ij},\phi_j);y_{ij},\Omega_{ij}),$$
with $\ell((\Theta_{ij},\phi);y_{ij},\Omega_{ij})=\log(p((y_{ij},\Omega_{ij});\Theta_{ij},\phi_j)), \: \forall i,j$.
In practice, the parameters vector $\phi$ is unknown but viewed as a nuisance parameter, since our main interest is the estimation of $\Theta$. To find an estimator $\hat{\Theta}$, we aim at maximizing the following penalized joint negative log-likelihood: 
\begin{equation}
\label{pb:MNARmodel}
(\hat{\Theta} , \hat{\phi}) \in \mathrm{argmin}_{\Theta,\phi} \ell(\Theta,\phi;y,\Omega) + \lambda \|\Theta\|_\star.
\end{equation}
It can be achieved using a Monte-Carlo Expectation Maximization  (MCEM) algorithm, whose two steps, iteratively proceeded, are given below:

\begin{itemize}
	\item \textbf{E-step:} the expectation (taking the distribution of the missing data given the observed data and the missing-data pattern) of the complete data likelihood is computed:
	\begin{multline}
	Q(\Theta,\phi|\hat{\Theta}^{(t)},\hat{\phi}^{(t)}) \\ =\mathbb{E}_{Y_\textrm{mis}}\left[\ell(\Theta,\phi;y,\Omega)|Y_\textrm{obs},M;\Theta=\hat{\Theta}^{(t)},\phi=\hat{\phi}^{(t)}\right]
	\end{multline}
	
	\item \textbf{M-step:} the parameters $\hat{\Theta}^{(t+1)}$ and $\hat{\phi}^{(t+1)}$ are determined as follows: 
	\begin{equation}
	\label{eqMstep}
	\hat{\Theta}^{(t+1)}, \hat{\phi}^{(t+1)} \in \textrm{argmin}_{\Theta,\phi} \: Q(\Theta,\phi|\hat{\Theta}^{(t)},\hat{\phi}^{(t)}) + \lambda\|\Theta\|_{\star}.
	\end{equation}
\end{itemize}

The E-step may be rewritten as follows:
\begin{multline*}
Q(\Theta,\phi |\hat{\Theta}^{(t)},\hat{\phi}^{(t)})=-\sum_{i=1}^{n}\sum_{j=1}^{p} C_1^{\Omega_{ij}} + C_2^{1-\Omega_{ij}}
\end{multline*}
where
\begin{align*}
C_1&= \log(p(y_{ij},\Omega_{ij};\Theta_{ij},\phi_j)) 
\\ C_2&=\int \log(p(y_{ij},\Omega_{ij};\Theta_{ij},\phi_j)) p(y_{ij}|\Omega_{ij};\hat{\Theta}_{ij}^{(t)},\hat{\phi}_j^{(t)})\mathrm{d}y_{ij}
\end{align*}
Note that the E-step is written as a sum of the E-steps for each $(i,j)$-th elements. If the $(i,j)$-th element is observed, we do not integrate and it leads to the first term; the second term corresponds to the missing elements.
By the lack of a closed form for $Q$, it is approximated by using a Monte Carlo approximation, denoted as $\hat{Q}$, $\forall i \in \left[1,n\right], \forall j \in \left[1,p\right]$:
\begin{multline*}
\hat{Q}_{ij}(\Theta,\phi|\hat{\Theta}^{(t)},\hat{\phi_j}^{(t)})= \\
-\frac{1}{N_s}\sum_{k=1}^{N_s}\log(p(v_{ij}^{k};\Theta_{ij}))+\log(p(\Omega_{ij}|v_{ij}^{k};\phi_j)), 
\end{multline*}
$\textrm{where }v_{ij}^k=\left\{
\begin{array}{ll}
y_{ij} \textrm{ if $\Omega_{ij}=1$,} \\
z_{ij}^{k} \textrm{ otherwise,}
\end{array}
\right.$
with $z_{ij}^{k}$ the realisation of $Z \sim p(y_{ij}|\Omega_{ij};\hat{\Theta}_{ij}^{(t)},\hat{\phi_j}^{(t)})$.

Note that $\hat{Q}$ is separable in the variables $\Theta$ and $\phi$, so that the maximization for the M-step may be independently performed for $\Theta$ and $\phi$:
\begin{align}
\label{max1}
\hat{\Theta}^{(t+1)} &\in \underset{\Theta}{\textrm{argmin}} \: \sum_{i=1}^{n} \sum_{j=1}^{p} \frac{1}{N_s}\sum_{k=1}^{N_s}-\log(p(v_{ij}^k;\Theta_{ij})) + \lambda\|\Theta\|_{\star} \\
\label{max2}
\hat{\phi}^{(t+1)} &\in  \underset{\phi}{\textrm{argmin}}  \: \sum_{i=1}^{n} \sum_{j=1}^{p} \frac{1}{N_s}\sum_{k=1}^{N_s} -\log(p(\Omega_{ij}|v_{ij}^k;\phi_j)).
\end{align}

Classical algorithms can be used: 
(accelerated) proximal gradient method to solve \eqref{max1} and the Newton-Raphson algorithm to solve \eqref{max2}. 

Moreover, for all $i \in \{1,\dots,n\}$ and $j \in \{1,\dots,p\}$ such that $y_{ij}$ is missing, we suggest the use of the sampling importance resampling (SIR) algorithm \citep{gordon1993novel} to simulate the variable $z_{ij}^k$. The detail is given in Appendix \ref{sec:SIR} and we take as a proposal distribution a Gaussian distribution.

\subsection{Adding the mask}
\label{sec:mask}

We now propose to directly include the information of the mask while considering the criterion \eqref{OptimMAR}, without explicitly modelling the mechanism, so that the new optimisation problem is written as follows:
\begin{equation}\label{optimMask}
\hat{\Theta} \in  \textrm{argmin}_{\Theta} \frac{1}{2} \left\| [\Omega \odot Y |\Omega] - [\Omega|\mathbf{1}] \odot [\Theta|\Omega] \right\|^2_F + \lambda \|\Theta\|_{\star},
\end{equation}
where $\mathbf{1} \in \mathbb{R}^{n\times p}$ denotes the matrix such that all its elements are equal to 1, and $[X_1|X_2]$ denotes the column-concatenation of matrices $X_1$ and $X_2$.
To solve \eqref{optimMask},  we could use again classical algorithms such as the (accelerated) iterative (SVD) soft-thresholding algorithm (Section \ref{sec:FISTA}). However, this approach does not take into account that the mask is made of binary variables and suggests  that the concatenated matrix $[Y \odot \Omega,\Omega]$ is Gaussian. Consequently, a better approach is to take into account the mask binary type by using  the low-rank model but extended to the exponential family. There is a vast literature on how to deal with mixed matrices (containing categorical, real and discrete variables) in the low-rank model, see for example \cite{udell2016generalized,liu2018pca,cai2013max}. \cite{robin2018main} suggested such a method, by using a data-fitting term based on heterogeneous
exponential family quasi-likelihood with a nuclear norm penalization:
\begin{equation}\label{optimMimi}
\hat{\Theta} \in  \textrm{argmin}_{\Theta} \sum_{i=1}^{n}\sum_{j=1}^{p} \Omega_{ij}\left( Y_{ij}\Theta_{ij}+g_j(\Theta_{ij})\right) + \lambda\|\Theta\|_{\star},
\end{equation}
where $g_j$ is a link function chosen according to the type of the variable $j$.
In our case, it allows to model the joint distribution of the concatenated matrix $[Y \odot \Omega, \Omega]$ of size $n \times 2p$ as follows  : (i) the data are assumed to be Gaussian, i.e. for all $j \in [1,p]$, $g_j(x)=\frac{x^2\sigma^2}{2}$ (ii) the missing-data pattern can be modelled by the Bernoulli distribution with success probability $1/(1+\exp(-\Theta_{ij}))$, i.e. for all $j \in [p+1,2p]$, $g_j(x)=\log(1+\exp(x))$. 
To solve \eqref{optimMimi}, a Penalized Iteratively Reweighted Least Squares algorithm called \texttt{mimi} (see \cite[page 12]{robin2018main}) is used. The advantage of such a strategy is to better incorporate the mask as binary features but this comes at a price of a more involved algorithm in comparison to \eqref{optimMask}. 

\subsection{FISTA algorithm}
\label{sec:FISTA}

To solve \eqref{OptimMAR}, \eqref{max1} and \eqref{optimMask} we suggest to use the FISTA algorithm, introduced by \cite{beck2009fast}, detailed in Appendix \ref{sec:FISTAalgo}, which corresponds to an accelerated version of the proximal gradient method. The acceleration is performed via momentum. The key advantage is that it converges to a minimizer at the rate of $\mathcal{O}(1/K^2)$ ($K$ is the number of iterations) in the case of $L$-smooth functions. 

This algorithm is of interest compared to the the non-accelerated proximal gradient method, that is shown in Appendix \ref{sec:Proxsoft} to be implemented in \texttt{softImpute-SVD} in the R package \texttt{softImpute} (see \cite{ softManual}): it is  known to converge only to the rate $O(1/K)$ \citep[Theorem 3.1]{beck2009fast}. To be more precise, another algorithm has been suggested that uses alternating least-squares \citep{hastie2015matrix} and departs from the previous one by solving a non-convex problem: it relies on the maximum margin matrix factorization approach 
(combined  with a final SVD thresholding). Therefore, although appealing numerically, the  algorithm known as \texttt{softImpute-ALS} is proven to converge only to a stationary point. 

\section{Simulations}
\label{Section3}
The parameter $\Theta$ is generated as a low-rank matrix of size $n \times p$ with a fixed rank $r < \min(n,p)$. The results are presented for $N$ simulations, for each of them: (i)  a noisy version $Y$ of $\Theta$ is considered,
$$Y = \Theta + \epsilon,$$
where $\epsilon$ is a Gaussian noise matrix with i.i.d.\ centered entries of variance $\sigma^2$, (ii)  MNAR missing values are introduced using a logistic regression, resulting in a mask $\Omega$ and (iii) only knowing $Y\odot \Omega$, we apply different methods to denoise and impute $Y$: 

\begin{enumerate}[label=(\alph*)]
	\item  \label{method:model} Explicit method (Model): in order to take into account the missing mechanism modelling, we apply the MCEM  algorithm to solve \eqref{pb:MNARmodel}, as detailed in Section \ref{sec:model};  note that either FISTA or \texttt{softImpute} are performed in the M-step. 
	\item \label{method:implicit}  Implicit method (Mask): the missing mechanism is implicitly integrated by concatenating the mask to the data, as detailed in Section \ref{sec:mask}.  When the binary type of the mask is neglected, FISTA or \texttt{softImpute} are used to solve \eqref{optimMask}. When taking into account the binary type of the mask, solving \eqref{optimMimi} is done by \texttt{mimi}.		
	\item \label{method:MAR}  MAR methods:  they consist in classical methods for low-rank matrix completion, proved to be efficient under the MCAR or MAR assumption,  and that aim at minimizing \eqref{OptimMAR}. The missing values mechanism is then ignored. They  encompass FISTA and \texttt{softImpute}. 
\end{enumerate}
We also include in \ref{method:implicit} and \ref{method:MAR}
the regularised iterative PCA algorithm \citep{verbanck2015regularised, josse2016denoiser} which uses another penalty than the nuclear norm one. We also compare all the methods to  the naive imputation by the mean (the estimation of $\Theta$ is obtained by replacing all values by the mean of the column).	 We performed an extended simulation study and  other more heuristic methods have been tested, such as the FAMD and MFA algorithms dedicated to mixed data or blocks of variables \citep{audigier2016principal} but they are not included in the article to make the plots more readable as the results were never convincing. The results presented are  representative of all the results obtained.

The results are presented for different matrix dimensions and ranks, mechanisms of missing values (MAR and MNAR), and percentages of missing data.	
The code to reproduce all the simulations is available on github \url{https://github.com/AudeSportisse/stat}.

\paragraph{Measuring the performance} To measure the methods performance, two types of normalized mean square errors (MSE) are considered: 
\begin{align}
\label{predErr}
&\mathbb{E}\left[\left\| (\hat{\Theta}-Y) \odot (1-\Omega)\right\|_F^2\right] \Biggm/ \mathbb{E}\left[\left\| Y \odot (1-\Omega) \right\|_F^2\right] \\
\label{totErr}
&\mathbb{E}\left[\left\| \hat{\Theta}-\Theta \right\|_F^2\right]  \Biggm/ \mathbb{E}\left[\left\| \Theta \right\|_F^2\right],
\end{align} 
that are respectively the prediction error, corresponding to the error committed when we impute values,  and the total error,  encompassing the prediction and the estimation error.

Some practical details on the algorithms are provided in the following paragraphs.
\paragraph{EM algorithm}
The stopping criterion used in the EM algorithm is the following: $$\frac{\|\hat{\Theta}^{(t)} - \hat{\Theta}^{(t-1)}\|_F}{\|\hat{\Theta}^{(t-1)}\|_F+\delta}\leq\tau,$$ where $\delta=10^{-3}$ and $\tau=10^{-2}$\footnote{Once the stopping criterion is met, $T=10$ extra iterations are performed to assure the convergence stability.}.
In addition, the E-step is performed with $N_s=1000$ Monte Carlo iterations. The key issue of this method is the run-time complexity largely due to this Monte Carlo approximation.

\paragraph{Tuning the algorithms hyperparameters}
When considering \eqref{OptimMAR}, \eqref{optimMask} and \eqref{eqMstep}, the regularisation parameter $\lambda$ is chosen among some fixed grid $\mathcal{G}=\{\lambda_1,\dots,\lambda_M\}$ to minimize either the prediction or the total errors. 
In the regularised iterative PCA algorithm, the hyper-parameter is the number of components to perform PCA, which can be found using cross-validation criteria. 
In the simulations, the noise level is assumed to be known. To overcome this hypothesis, one can use standard estimators of the noise level such as the ones of \cite{gavish2017optimal} and \cite{josse2016denoiser}.

\subsection{Univariate missing data}
\label{sec:univ}
\begin{figure}
	\begin{center}
		\includegraphics[width=0.5\textwidth]{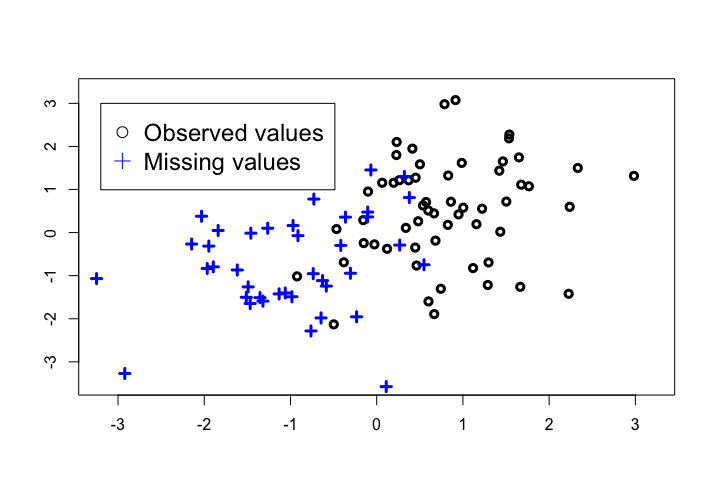}
		\caption{ \label{fig:data} Introduction of MNAR missing values using a logistic regression \eqref{eq:log_reg}, with $\phi_1=3$ and $\phi_2=0$.  One can see that the the highest values of $y_{i1}$ are missing, mimicking a cutoff effect.}
	\end{center}
\end{figure}

\begin{figure}
	\begin{bigcenter}
		\includegraphics[width=1\textwidth]{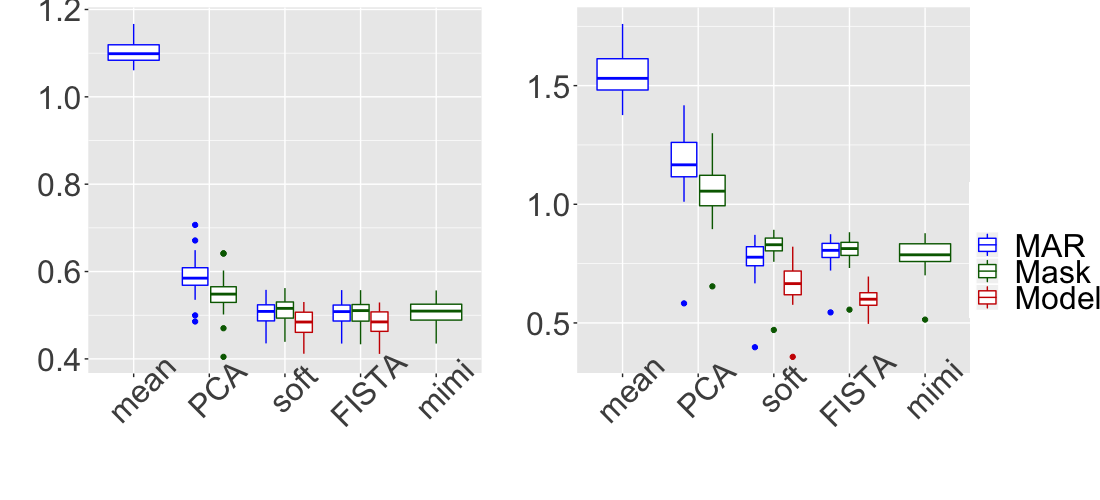}
		\caption{ \label{fig:Univariate} Univariate missing data: total error (left) and prediction error (right) for the methods \ref{method:model} in red, \ref{method:implicit} in green and \ref{method:MAR} in blue. }
	\end{bigcenter}
	
\end{figure}

Let us consider  a simple case with $n=100$ and $p=4$, the rank of the parameter matrix is $r=1$ and $\sigma^2=0.8$. Assume that only one variable has missing entries.
The missing values are introduced by using the self-masked MNAR mechanism. The missingness probabilities are then given as follows:
\begin{equation}
\label{eq:log_reg}
\forall i \in [1:n], p(\Omega_{i1}=0|y_{i1}; \phi)=\frac{1}{1+e^{-\phi_1(y_{i1}-\phi_2)}}
\end{equation}
The parameters of  the logistic regression are chosen to mimic a cutoff effect, see Figure \ref{fig:data}. Indeed, extrapolating imputed values can be challenging and classical methods are expected to introduce a large prediction bias. 
Given the previous parameters choice, the percentage of missing values is $50\%$ in expectation for the missing variable,  corresponding to $12.5\%$ missing values in the whole matrix. In Figure \ref{fig:Univariate}, the three methods \ref{method:model}, \ref{method:implicit} and \ref{method:MAR} are compared in such a setting, using boxplots on MSE errors for $N=50$ simulations. In this MNAR setting, the proposed model-based method \ref{method:model}, in red in Figure \ref{fig:Univariate},  aiming at minimizing \eqref{pb:MNARmodel} -specially designed for such a setting- gives better results glo\-bal\-ly for the total error with a significant improvement on the prediction of missing values (either when FISTA or \texttt{softImpute} is used in the M-step of the MCEM algorithm). 

In addition, the implicit methods 
\ref{method:implicit}, in green in Figure \ref{fig:Univariate}, working on the concatenation of the mask and the data, either based on a binomial modeling of the mechanism (\texttt{mimi}, solving \eqref{optimMimi}), or neglecting the binary feature of the mask (FISTA and \texttt{softImpute}, solving \eqref{optimMask}), do not lead to improved performance compared to the MAR method \ref{method:MAR} (FISTA and \texttt{softImpute})  in terms of prediction or estimation errors. 
On the contrary, the implicit method \ref{method:implicit} working on the concatenation of the mask and the data, based now on the regularized iterative PCA improves both estimation and prediction errors compared to the regular PCA algorithm used in the MAR method \ref{method:MAR}.  However the obtained prediction error does not compete with performance of regular MAR completion algorithms (FISTA and \texttt{softImpute}).

Note also that the results of both SVD algorithms, \texttt{softImpute} and FISTA, are similar in terms of estimation and prediction error, but FISTA has the advantage to improve the numerical convergence to a minimizer. 

In conclusion on the univariate case, (i) modelling the missing me\-cha\-nism outperforms any other method, particularly in terms of prediction error; (ii) implicit methods \ref{method:implicit} have limited interest, except to improve the regular PCA algorithm.

\subsection{Bivariate missing data}
\label{Bivar}

We consider now a higher dimensional case: $n=100$ and $p=50$ and the rank of the parameter matrix is $r=4$. The noise level is $\sigma^2=0.8$, as in Section \eqref{sec:univ}. The missing values are introduced on two variables by using the following MNAR mechanism, for all $ i \in [1,n]$ and $j \in [1,2]$,
$$p(\Omega_{ij}=0|y_{ij}; \phi)=\frac{1}{1+e^{-\phi_{1j}(y_{ij}-\phi_{2j})}}$$
\[\textrm{ where}
\left\{
\begin{array}{ll}
\phi_{1j}=3 , \phi_{2j}=0 \textrm{ if $j=1$}, \\
\phi_{1j}=2 , \phi_{2j}=1 \textrm{ if $j=2$}.
\end{array}
\right. \]

\begin{figure}
	\begin{bigcenter}
		\includegraphics[width=1\textwidth]{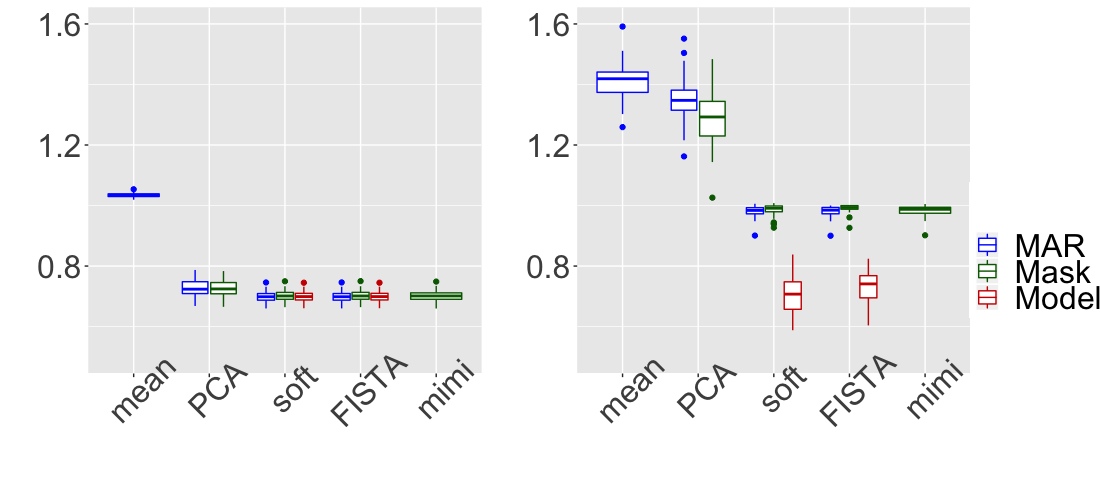}
		\caption{ \label{fig:bivariateMNAR} Bivariate missing data: total error (left) and prediction error (right) for the methods \ref{method:model} in red, \ref{method:implicit} in green and \ref{method:MAR} in blue.}
	\end{bigcenter}
\end{figure}

This parameters choice leads to $50\%$ missing values in $Y_{.1}$ and $20\%$ in $Y_{.2}$ mimicking a cutoff effect again. In Figure \ref{fig:bivariateMNAR}, the methods \ref{method:model}, \ref{method:implicit} and \ref{method:MAR} are compared in such a setting, using boxplots on MSE errors for $N=50$ simulations. 

The model-based method \ref{method:model}, designed for the MNAR setting, give significant better results than any other method  in terms of prediction error. 
The mask-adding methods \ref{method:implicit} lead to no significant improvement compared to classical MAR methods,
either by solving \eqref{optimMask} using FISTA, \texttt{softImpute}, or solving \eqref{optimMimi} via \texttt{mimi}.
One can note that the PCA algorithm still benefits from the concatenation with the mask in terms of prediction error,  but to a lesser extent  than in the univariate case. 

Overall, the poor performance of the mask-adding methods \ref{method:implicit} can be explained 
by the dimensionality issue and the small weight of the added mask variables. Indeed, in this higher dimensional case with bivariate missing variables, only  two informative binary variables corresponding to the mask are really concatenated to a 50-column matrix. 

Note that in terms of total error, the advantages of model-based methods \ref{method:model} are no longer visible, which can be explained by the very low percentage of missing data ($1.5\%$) (see Section \ref{MoreMissing} in which more missing values are considered).

\subsection{Multivariate missing data}
\label{MoreMissing}

\begin{figure}
	 \vspace*{-0.15\textwidth}
	\begin{bigcenter}
		\includegraphics[width=1.1\textwidth]{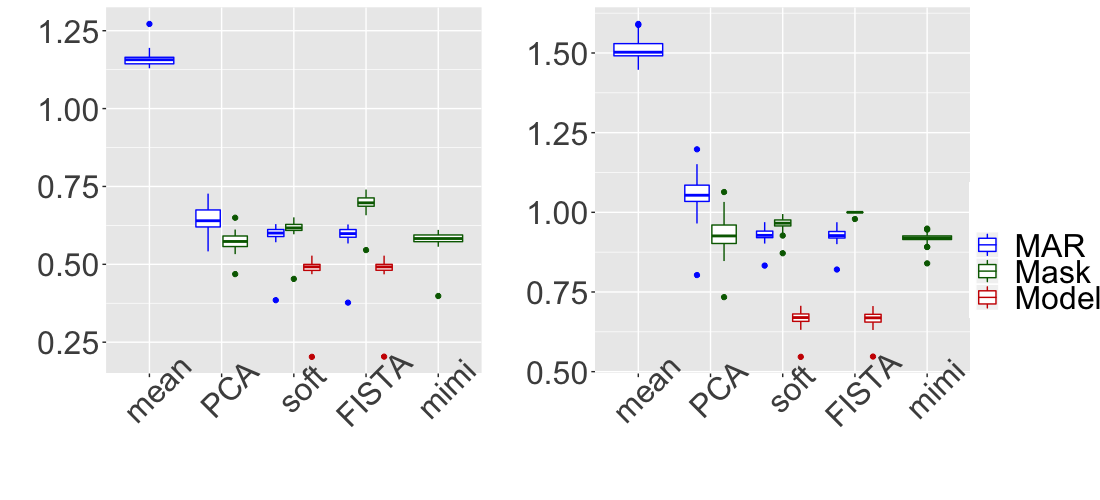}
		\includegraphics[width=1.1\textwidth]{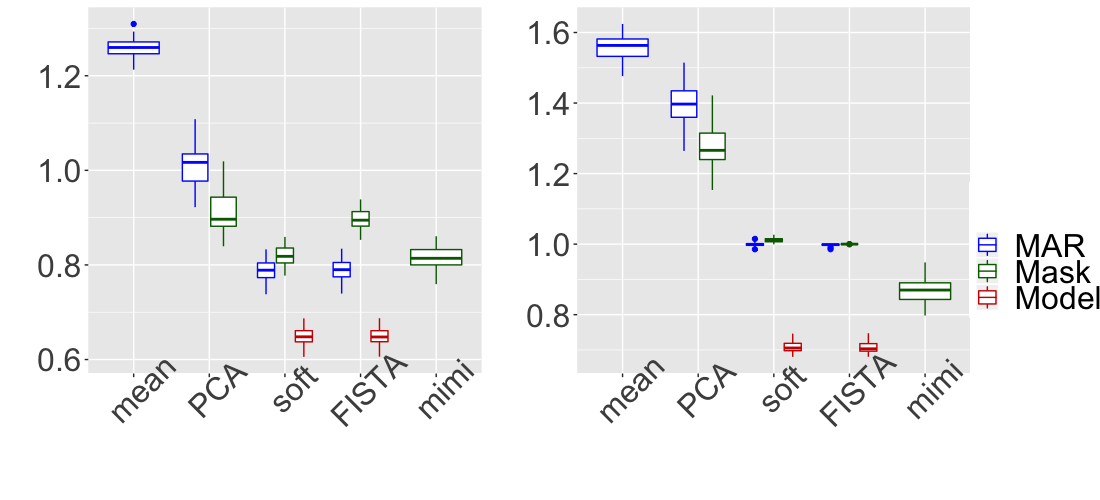}
		\includegraphics[width=1.1\textwidth]{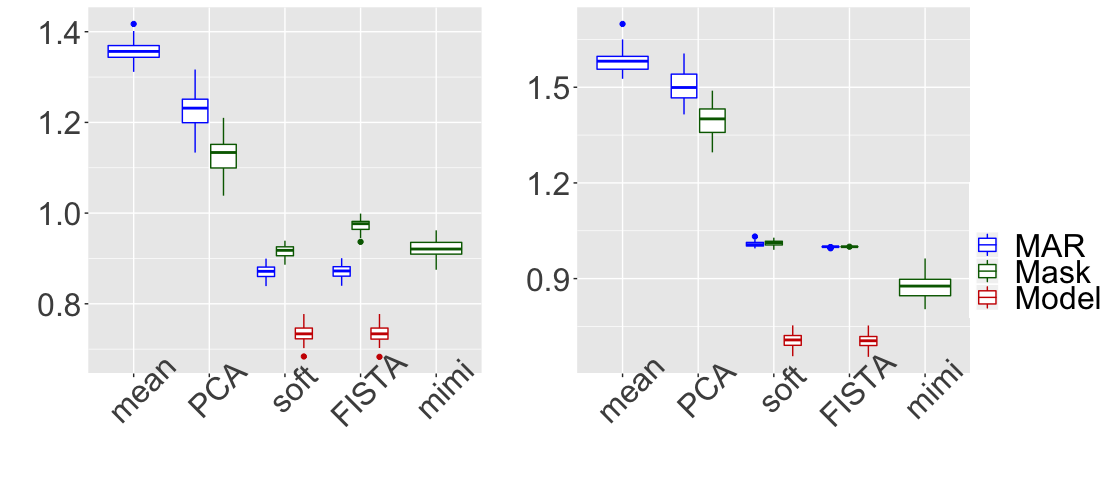}
		\caption{\label{Fig5} Multivariate MNAR missing data: total error (left) and prediction error (right) for the methods \ref{method:model} in red, \ref{method:implicit} in green and \ref{method:MAR}. Three noise settings are considered: on top strong signal ($\sigma^2=0.2$),  middle noisy data ($\sigma^2=0.5$), bottom very noisy data ($\sigma^2=0.8$).{}}
	\end{bigcenter}
\end{figure}

We consider now a multivariate missing data case for the following dimensional setting: $n=100$, $p=20$ and $r=4$. The missing values are introduced on ten variables by using the following MNAR mechanism, for all  $i,j \in [1,n]\times[1,10]$, 
$$p(\Omega_{ij}=0|y_{ij}; \phi)=\frac{1}{1+e^{-\phi_{1}(y_{ij}-\phi_{2})}}.$$
Note that the parameters of the missingness mechanism are the same for each element, this can be easily extended to a more general case. The parameters choice leads to $25\%$ missing values in the whole matrix. The results are presented in Figure \ref{Fig5} for $N=50$ simulations  and different noise levels, $\sigma^2 = 0.2,0.5$ or $0.8$. 

First, one can note that the model-based method \ref{method:model} provides the best result both in estimation and prediction error regardless the noise level (and whatever FISTA or \texttt{softImpute} used in the MCEM). Of course, this performance improvement comes at the price of a computational cost due to the Monte Carlo approximations needed in the MCEM algorithm.

Regarding the implicit methods \ref{method:implicit}, 
the mask-adding techniques handling the concatenation of the data and the mask matrix as Gaussian (FISTA and \texttt{softImpute}) miss to improve both estimation and prediction errors compared to their MAR version.
However, the variant \texttt{mimi} modelling the mask with a binomial distribution always largely outperforms MAR methods \ref{method:MAR} in terms of prediction (while the improvement in terms of estimation error is only visible at a low noise level). Therefore, the mask-adding approach can implicitly capture the MNAR missing mechanism, when the mask is really considered as a matrix of binary variables. This comes at the price of a more involved algorithm $\texttt{mimi}$ able to take into account mixed variables, but that remains far less computationally expensive than the model-based approach. Indeed, for an estimation/prediction of one parameter matrix $\Theta$, the process time for a computer with a processor Intel Core i5 of 2,3 GHz is 0.0549 seconds for the MAR method with \texttt{softImpute}, 3.215 seconds for the implicit method with \texttt{mimi} and 13.069 minutes for the model-based method with \texttt{softImpute} when $50\%$ of the variables are missing.

As a side comment, in this high-dimensional setting, one can note that the PCA algorithm still benefits from adding the mask,  which is a variant of method \ref{method:implicit},  compared to the regular PCA method, both in estimation and prediction error. However the mask-adding PCA algorithm only compete the mask-adding methods based on iterative SVD thresholding (FISTA, \texttt{softImpute}) at a low noise level. 

\subsection{Sensitivity to model misspecifications}	


\paragraph{Deviation in the missing-data mechanism setting}

Here, the missing values are introduced by using the MAR mechanism. It allows to test the stability of model-based methods, designed for the MNAR setting,  to a deviation in the missing mechanism. The missingness probabilities are given as follows in such a setting: 
\begin{align}
\label{eq:missing_MAR}
\forall i \in [1,n], p(\Omega_{i1}=0|y_{i2}; \phi)=\frac{1}{1+e^{-\phi_1(y_{i2}-\phi_2)}},
\end{align}
meaning that the probability to have a missing value in $Y_1$ depends on the value of $Y_2$.

First, let us consider the setting of Section \ref{sec:univ}, i.e. $n=100$, $p=4$, $r=1$. 

\begin{figure}
	\begin{bigcenter}
		\includegraphics[width=1\textwidth]{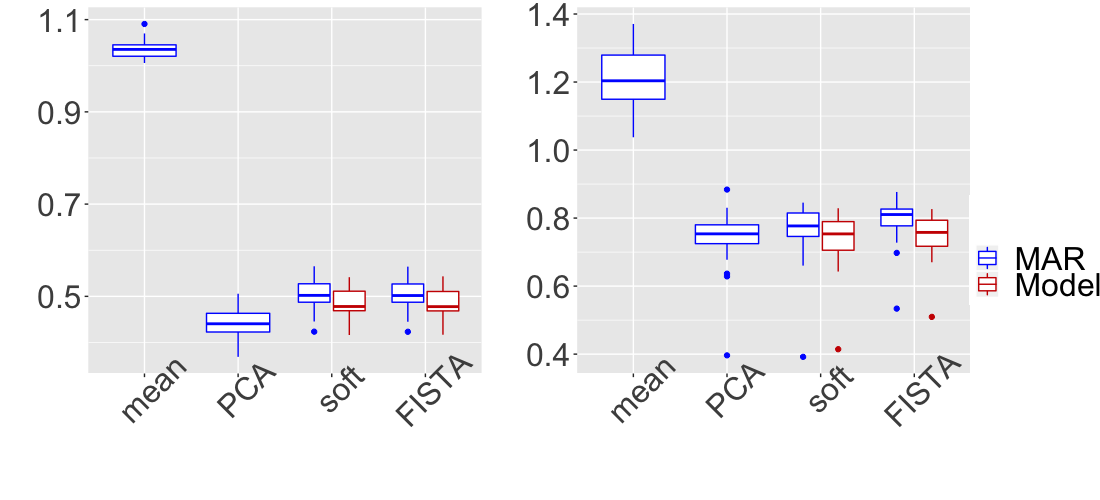}
		\caption{\label{fig:univariateMAR} Comparison of methods performance when the missing data are of type MAR (for $N=50$ simulations) with a rank one: total error (left) and prediction error (right) for different methods and algorithms.}
	\end{bigcenter}
\end{figure}

In Figure \ref{fig:univariateMAR}, we observe that the model-based method \ref{method:model} improves both the estimation and the prediction, which is not expected in a MAR setting.  However, this can be explained because 
of the rank is one  which implies that there are only small differences between MNAR and MAR (the second variable's value is directly linked to the missing one's value). Consequently, modelling a MNAR mechanism is enough to retrieve information on such a MAR missing mechanism.

\begin{figure}
	\begin{bigcenter}
		\includegraphics[width=1\textwidth]{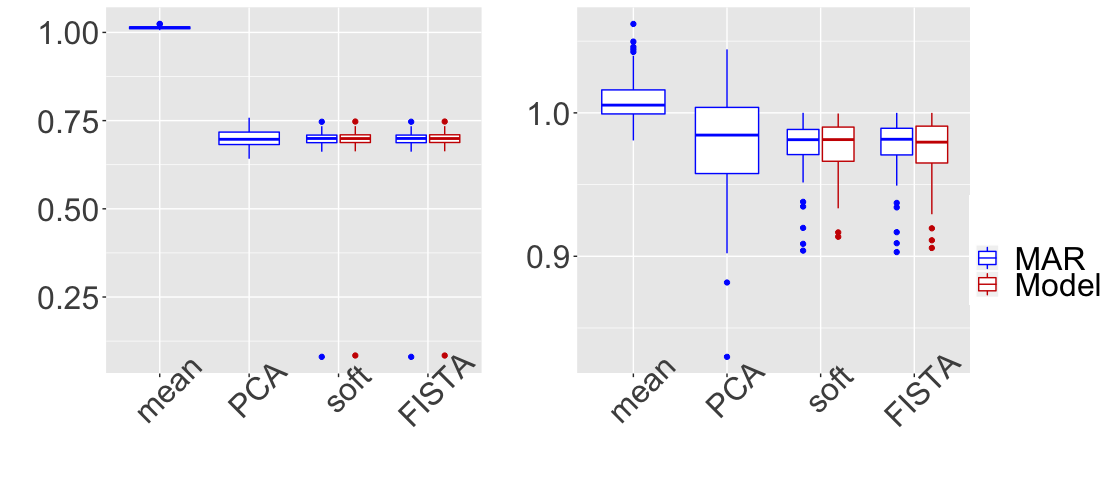}
		\caption{\label{fig:bivariateMAR} Comparison of methods performance when the missing data are of type MAR (for $N=50$ simulations) with a rank four (the MAR mechanism depends on a decorrelated variable to the missing one): total error (left) and prediction error (right) for different methods and algorithms.}
	\end{bigcenter}
\end{figure}

To avoid this case, we consider the setting of Section \ref{Bivar}, i.e.\ $n=100$, $p=50$, $r=4$, with a MAR missing mechanism as described by \eqref{eq:missing_MAR}, however, the second variable involved is chosen to be decorrelated from the missing one (which is possible given the rank is $4$). In such a case, there is no equivalence between the missing values that are simulated to be MAR and the mechanism we model as MNAR. Figure \ref{fig:bivariateMAR} shows that the model-based approach does not lead to any improvement compared to regular methods used for MAR methods; but more importantly, it does not degrade the results either which highlights the robustness of the approach with respect to deviations from the model.

\begin{figure}
	\begin{bigcenter}
		\includegraphics[width=1\textwidth]{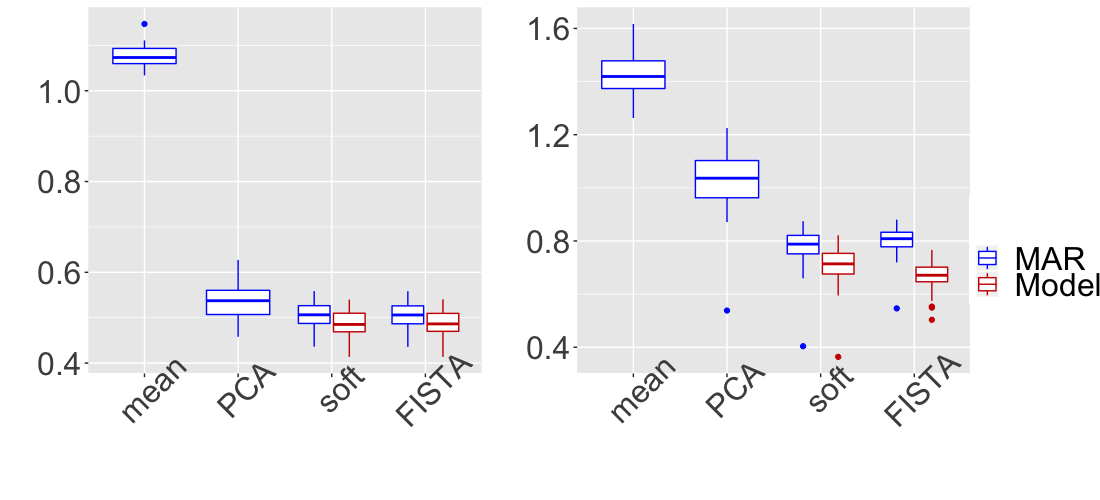}
		\caption{\label{fig:univariateprobit}  Univariate MNAR missing data parametrized with a probit model  for $N=50$ simulations: total error (left) and prediction error (right) for different methods and algorithms. Note that the methods modeling the missing mechanisms use the logistic model. }
	\end{bigcenter}
\end{figure}

\paragraph{Deviation in the logistic regression setting} We now want to test the robustness of our model-based method \ref{method:model} to a misspecification of the logistic model, given by \eqref{densmecha}. To do so, missing values are introduced by a MNAR missing-data mechanism based on the following probit model, the missingness probabilities are then:
$$
\forall i \in [1,n], \qquad p(\Omega_{i1}=0|y_{i1}; \phi)=F(y_{i1}),$$
where $F$ is the quantile function the standard Gaussian cumulative distribution function. Consider the setting of Section \ref{sec:univ}, i.e. $n=100$, $p=4$, $r=1$.
In Figure \ref{fig:univariateprobit}, we observe that the model-based methods \ref{method:model} globally improves the results for both errors \eqref{totErr} and \eqref{predErr}. 
Very similar results to the ones of Section \ref{sec:univ} are obtained, meaning that the model-based method \ref{method:model} behaves well to a deviation of the logistic regression modelling. 

\section{Application to clinical data} \label{sec:Traumabase}

\subsection{Motivation}

Our work is motivated by a public health application with APHP TraumaBase$^{\mbox{\normalsize{\textregistered}}}$\footnote{http://www.traumabase.eu/} Group (Assistance
Publique - Hopitaux de Paris) on the management of
traumatized patients. Major trauma, i.e. injuries that endanger a person's life or functional integrity, have been qualified as a worldwide public health challenge and a major source of mortality (first cause in the age group 16-45) in the world by the WHO \citep{hay2017global}. Hemorrhagic shock and traumatic brain injury have been identified as the lead causes of death. Effective and timely management of trauma is crucial to improve outcomes, as delays or errors entail high risks for the patient. 




\subsection{Data description}

A subset of the trauma registry containing the clinical measurements of $3168$ patients with brain trauma injury is first selected. 


Our aim is to predict from pre-hospital measurements whether or not the tranexomic acid\footnote{the tranexomic acid is an antifibrinolyic agent which reduces blood loss.} should be administrated on arrival at the hospital. In the dataset, the variable \textit{Tranexomic.acid} is the decision made by the doctors, which is considered as ground truth. This variable is equal to $1$ if the doctors have decided to administrate tranexomic acid, $0$ otherwise. 

\begin{figure}
	\begin{bigcenter}
		\includegraphics[width=1\textwidth]{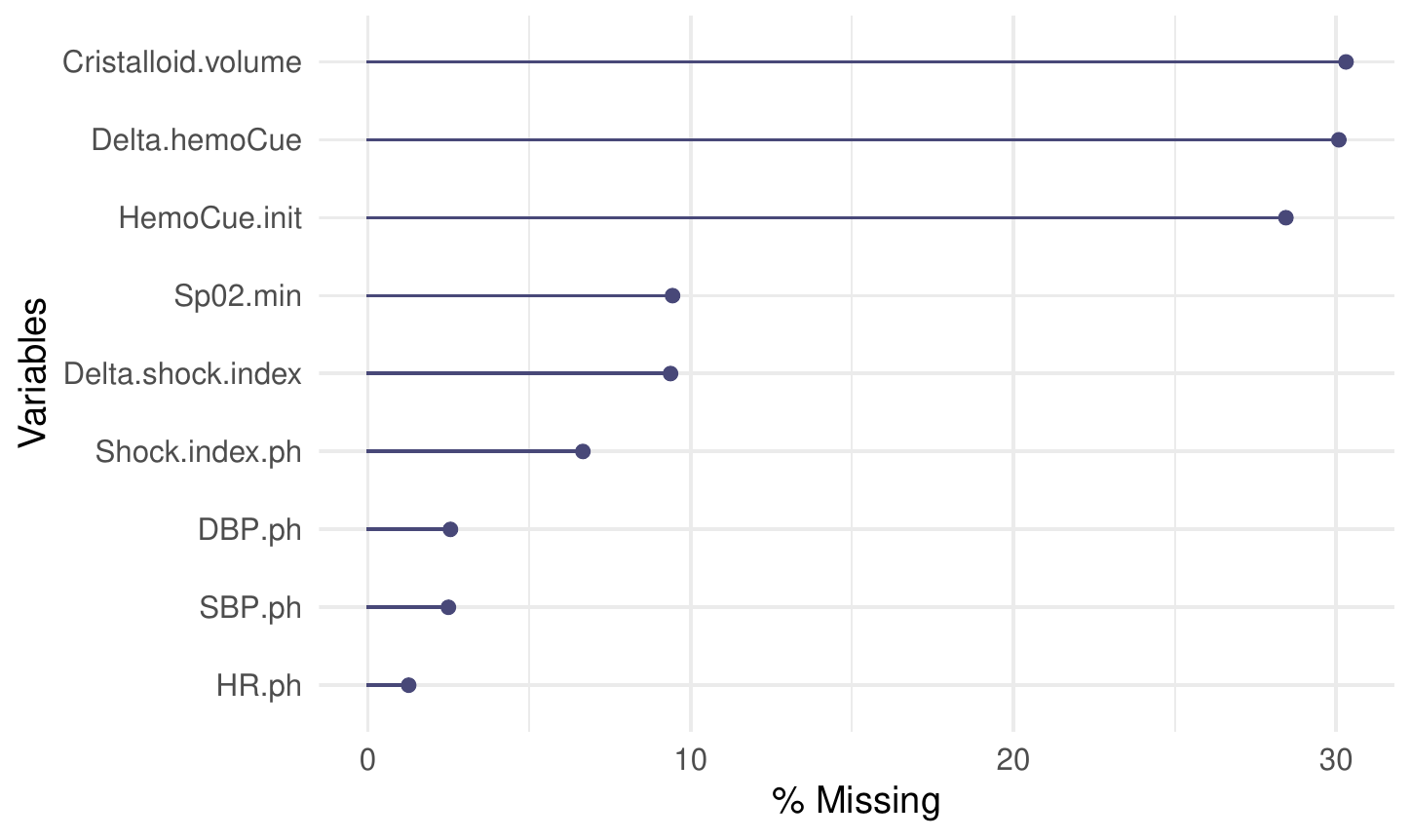}
		\caption{\label{fig:percentage} Percentage of missing values in each variable.}
	\end{bigcenter}
\end{figure}

Nine quantitative variables containing missing values are selected by doctors. In Figure \ref{fig:percentage}, one can see the percentage of missing values in each variable, varying from $1.5$ to $30\%$, leading to $11\%$ is the whole dataset. 
After discussion with doctors, almost all variables can be considered to have informative missingness. For example, when the patient's condition is too critical and therefore his heart rate (variable \textit{HR.ph}) is either high or low, the heart rate may not be measured, as doctors prefer to provide emergency care. The heart rate itself can then be qualified of self-masked MNAR, and the other variables, either of MNAR or MAR. Both percentage and nature of missing data demonstrate the importance of taking appropriate account of missing data. 
More information on the data can be found in Appendix \ref{sec:detailstrauma}.

In the following, two questions are addressed. Firstly, 
we compare the validity of the imputation methods in terms of prediction of the tranexomic acid administration based on the different imputed data.
Secondly,  we test the methods in terms of their imputation performance. 


\subsection{Prediction of administration of the tranexomic acid} 

We consider a two-step procedure:
\begin{itemize}
	\item \uline{Step 1: imputation of the explanatory variables.}
	As a pre\-pro\-ces\-sing step, we impute missing data in the explanatory variables, beforehand proceeding to the classification training.  
	Imputation is performed using the model-based method \ref{method:model}, the implicit methods \ref{method:implicit} or the MAR methods  \ref{method:MAR}. 
	All these methods are compared to the naive imputation by the mean. 
	\\
	\item \uline{Step 2: classification task} which consists in predicting the administration or not of the tranexomic acid. Therefore, we are looking for the prediction function $f$ such that 
	$$Z\simeq f(Y^{\text{imp}}),$$
	where $Z \in \{0,1\}^n$ is equal to 1 (resp. 0) if the tranexomic acid is (resp. not) administered, and $Y^{\text{imp}} \in \mathbb{R}^{n\times p}$ represents the nine imputed explanatory variables discussed above. 
	Based on these new-filled design matrices formed in Step 1, the classification is always done using either random forests or logistic regression.

	Since not administering tranexomic acid by mistake can be vital, for the training and testing errors, we use a dissymetrized loss function where the cost of false negatives is much more than of false positives as follows
	\begin{equation}\label{eq:validationerror}
	l(\hat{z},z)=\frac{1}{n}\sum_{i=1}^{n}w_0\mathrm{1}_{\{z_i=1,\hat{z}_i=0\}}+w_1\mathrm{1}_{\{z_i=0,\hat{z}_i=1\}},
	\end{equation}
	where $w_0$ and $w_1$ are the weights for the cost of false negative and false positive respectively, s.t. $w_0 + w_1 = 1$ and $\omega_0=5\omega_1$. 
\end{itemize}

The dataset is divided into training and test sets (random selection of $80-20\%$) and the prediction quality on the test set is compared according to different indicators such as the accuracy, the sensitivity,  etc.




Table \ref{table:predres} compares results when random forests are used as a prediction method. In this setting, mean imputation gives among the best results on all the metrics which is in agreement with recent results on its consistency when used with a powerful learner, see \citet{josse2019consistency}.  Nevertheless,  the model-based method \ref{method:model} is very competitive.  The proposed implicit methods result in the best performances in terms of the sensitivity which is particularly relevant for the application.

Table \ref{table:predres2} compares results  when the prediction is performed with logistic regression. For almost all criteria, and especially on sensitivity the model-based method \ref{method:model} leads to the best performances. The standard deviations are also smaller with the model based approach in comparison with the implicit methods.  

Therefore, the model-based method  performs well regardless of  the prediction method used.

\begin{table}
	\begin{center}
	\begin{tabular}{l|l|ll|ll|l}
		& {\color[HTML]{CB0000} Model} & \multicolumn{2}{l|}{{\color[HTML]{009901} Mask}} & \multicolumn{2}{l|}{{\color[HTML]{3531FF} MAR}}  &\\
		& {\color[HTML]{CB0000} soft} & {\color[HTML]{009901} mimi} & {\color[HTML]{009901} soft} & {\color[HTML]{3531FF} soft} & 
		{\color[HTML]{3531FF} PCA} & mean \\
		\hline
		error & \textbf{12.5} & 16.0 & 15.8 & 14.8 &13.6 & \textbf{13.0} \\ 
		sd & 3.3 & 2.8 & 4.9 & 5.0 & 3.2 & 2.1 \\ 
		\hline
		AUC         & \textbf{85.4} & 83.9 & 84.6 & 84.6 & \textbf{85.5} &  \textbf{85.2} \\ 
		sd & 1.6 & 1.7 & 1.8 & 2.0 & 1.4 & 2.2 \\ 
		\hline
		acc    & \textbf{79.5} & 77.8 & 77.6 &  78.6 & \textbf{79.9}& \textbf{80.7} \\ 
		sd & 5.0 & 3.2 & 5.0 & 5.2 & 3.4 & 3.1 \\ 
		\hline 
		pre   & \textbf{47.5} & 45.0 & 45.1 & 46.5 & 45.2 &  \textbf{48.7} \\  
		sd   & 6.7 & 4.2 & 8.2 & 8.3 & 5.9 & 5.0 \\ 
		\hline
		sen & 76.5 & \textbf{78.1} & \textbf{78.2}  & \textbf{77.4} & 72.4 & 76.0 \\ 
		sd & 6.1 & 3.4 & 5.7 & 5.4 & 3.2 & 4.5 \\ 
		\hline
		spe & 80.2 & 77.7 & 77.4 &  78.9 & \textbf{80.8} & \textbf{81.7} \\ 
		sd & 7.2 & 4.4 & 7.2 & 7.3 & 4.6 & 4.6 
	\end{tabular}
	\caption{\label{table:predres} \textbf{By using random forest for the classification.} Comparison of the mean of different prediction criteria over ten simulations  (values are multiplied by 100). 
		Error corresponds to the validation error with the loss described in \eqref{eq:validationerror}. AUC is the area under ROC; the accuracy (acc) is the number of true positive plus true negative divided by the total number of observations; the sensitivity (sen) is defined as the true positive rate; specificity (spe) as the true negative rate; the precision (pre) is the number of true positive over all positive predictions. The lines sd correspond to standard deviations. 
		The three best results are in bold.}
	\end{center}
\end{table}


\begin{table}
	\begin{center}
	\begin{tabular}{l|l|ll|ll|l}
		& {\color[HTML]{CB0000} Model} & \multicolumn{2}{l|}{{\color[HTML]{009901} Mask}} & \multicolumn{2}{l|}{{\color[HTML]{3531FF} MAR}}  &\\
		& {\color[HTML]{CB0000} soft} & {\color[HTML]{009901} mimi} & {\color[HTML]{009901} soft} & {\color[HTML]{3531FF} soft} & 
		{\color[HTML]{3531FF} PCA} & mean  \\
		\hline
		error & \textbf{13.5} & \textbf{13.3} & 15.5 & 15.5 & 13.8 & 13.7 \\ 
		sd & 2.4 & 4.5 & 3.9 & 3.9 & 3.3 & 2.1 \\ 
		\hline
		AUC         & \textbf{82.6} & 78.7 & 81.9 & 81.9 & 82.1 & 82.0 \\ 
		sd & 2.4 & 2.3 & 2.4 & 2.4 & 2.5 & 2.4 \\ 
		\hline
		acc    & \textbf{80.1} & 79.3 & 77.6 & 77.6 & 79.6 & 79.8 \\
		sd & 3.7 & 6.9 & 6.1 & 6.1 & 5.1 & 3.3 \\ 
		\hline
		pre & \textbf{47.7} & 46.2 & 47.0 & 46.0 & 45.1 & 46.9 \\ 
		sd & 4.1 & 7.9 & 6.4 & 5 & 5.2 & 3.2 \\ 
		\hline
		sen   & \textbf{74.8} & 67.0 & 73.7 & 73.8 & 73.7 & 73.9 \\
		sd   & 5.1 & 4.4 & 7.6 & 7.7 & 6.5 & 5.5 \\ 
		\hline
		spe & 81.3 & \textbf{82.0} & 78.4 & 81.1 & 81.0 &  78.4 \\ 
		sd & 3.7 & 3.6 & 6.1 & 6.2 & 5.1 &  3.3 
	\end{tabular}
	\caption{\label{table:predres2} \textbf{By using logistic regression for the classification.} Comparison of the mean of different prediction criteria over ten simulations  (values are multiplied by 100). 
		Error corresponds to the validation error with the loss described in \eqref{eq:validationerror}. AUC is the area under ROC; the accuracy (acc) is the number of true positive plus true negative divided by the total number of observations; the sensitivity (sen) is defined as the true positive rate; specificity (spe) as the true negative rate; the precision (pre) is the number of true positive over all positive predictions. The lines sd correspond to standard deviations. 
		The two best results are in bold.}
	\end{center}
\end{table}



\subsection{Imputation performances}	

\begin{figure}
	\begin{bigcenter}
		\includegraphics[width=0.8\textwidth]{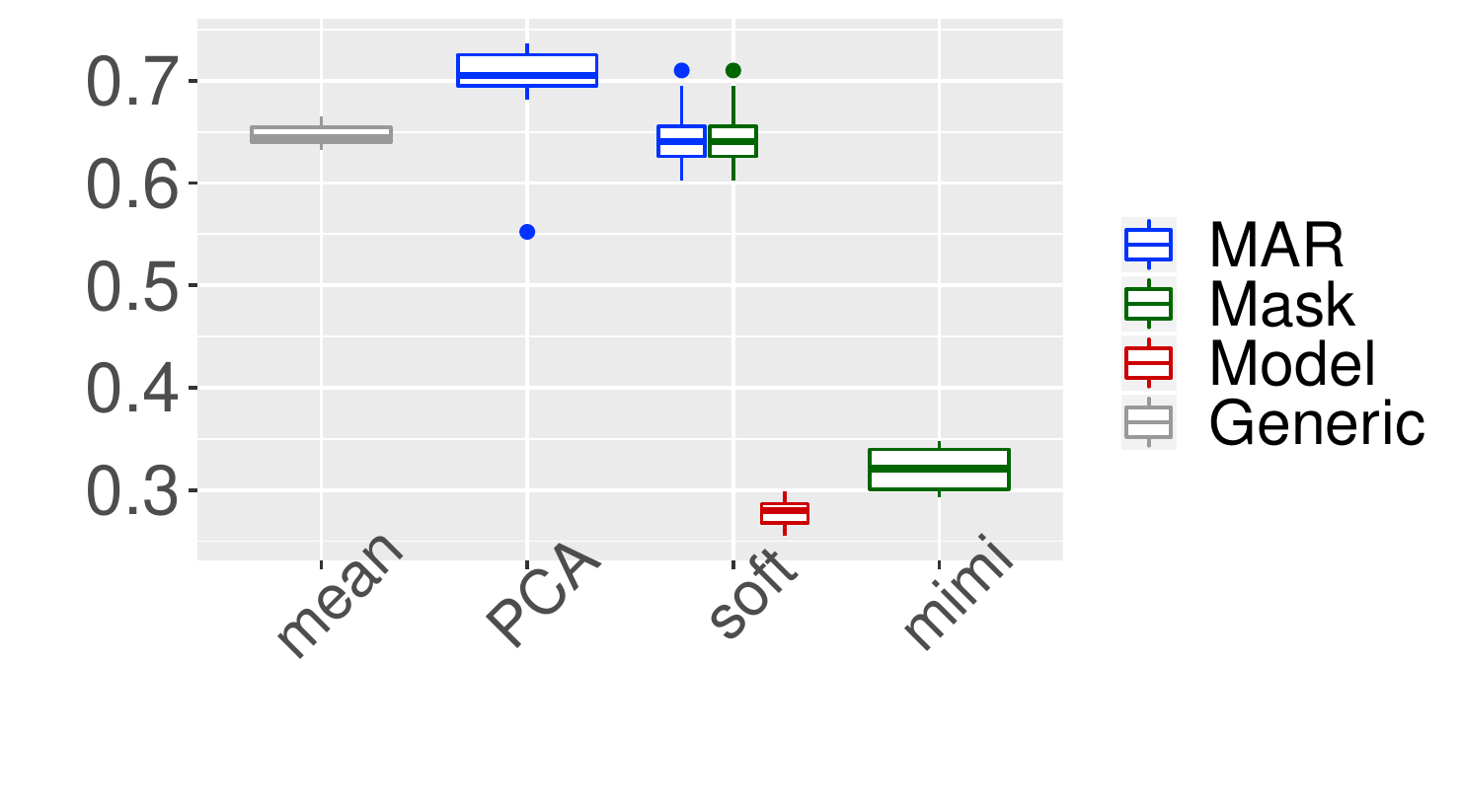}
		\caption{\label{fig:errorimp} Comparison of the imputation error (for ten simulations). 
		}
	\end{bigcenter}
\end{figure}

As the methods are initially designed for imputation, we perform simulations on the real dataset. In order to be able to measure the quality of the imputation,  some additional MNAR values are introduced in the variable \textit{Shock.index.ph},  which is a variable with MNAR missing values (according to  doctors) that contains initially $7\%$ of missing values.
The missing values are introduced by using the self-masked mechanism described in \eqref{eq:log_reg}. The choice of parameters in the logistic regression leads to 35\% missing values.
In the model-based method \ref{method:model}, the variables are scaled before each EM iteration to give the same weight to each variable. 
Besides, the noise level $\sigma^2$ is estimated using
the residual sum of squares divided by the number of observations
minus the number of estimated parameters as suggested in \cite{josse2016denoiser}, 
$$\hat{\sigma^2}=\frac{\|Y-\sum_{l=1}^{r}u_ld_lv_l\|^2_2}{np-nr-rp+r^2},$$
where $u_l$, $v_l$ and $d_l$ are the singular vectors and the singular values from the singular value decomposition of $Y$. We let $r$ denote the rank of $Y$, estimated here using cross-validation \citep{josse2012selecting}.
In Figure \ref{fig:errorimp}, the three methods \ref{method:model}, \ref{method:implicit} and \ref{method:MAR} are compared using boxplots of the prediction error over ten simulations.
The proposed method \ref{method:model}, designed for the MNAR setting, gives significantly smaller prediction error than other methods. 
Besides, the other proposed methods \ref{method:implicit}, taking the mask into account, also improve prediction errors compared to the classical MAR methods \ref{method:MAR}. 

\section{Discussion} 
\label{sec:discussion}

In this article two methods have been suggested for  handling self-masked MNAR data in the low-rank context: explicit modeling of the mechanism or implicit consideration by adding the mask. The first method is clearly the most successful in terms of prediction or estimation errors. Moreover, it is robust to model misspecifications. However, one should note that, on the one hand it can be computationally expensive, and then hardly scalable in the high-dimensional multivariate missing setting and on the other hand, it is a parametric approach. Therefore, the implicit method handling both the data and the mask matrices, when taking into account the binary distribution of the latter, may be regarded as the right alternative. Both methods can handle MNAR and MAR data simultaneously.

As a take-home message, one should keep in mind that (i) if there are a few missing variables, the model-based method is extremely relevant; and (ii) when many variables can be missing, the implicit method, that models the mask using a binomial distribution, has empirically proven to provide better imputation. 


Note that the logistic regression assumption may seem restrictive but the proposed approach could be easily adapted to other distributions such as the probit one. 

We pointed out that when the rank is one, there are few differences between MAR and MNAR, which implies that MNAR missing values could be handled without specifiying a model. This is in  line with the work of  \cite{mohan2018estimation} in regression using graphical models and it would be interesting to extend their work to low-rank models.

As directions of future research, one could also extend this work to data matrices containing mixed variables (quantitative and categorical variables) with MNAR data, so that the logistic regression model should include the case of categorical explanatory and output variables. 

In addition, in this paper, we focus on single imputation techniques where a unique value is predicted for each missing value. Consequently, it can not reflect the variance of prediction. It would be very interesting to derive confidence intervals for the predicted value, for instance by considering multiple imputation methods \citep{rubin2004multiple}.





\section*{Acknowledgments}
The authors are thankful for fruitful discussion with Fran\c cois Husson, Wei Jiang, Imke Mayer and Geneviève Robin. 

\begin{appendices}
	\section{The FISTA algorithm} 
	\label{sec:FISTAalgo}
	We first present the proximal gradient method. The following optimisation problem is considered:
	$$\hat{\Theta} \in \textrm{argmin}_\Theta \: h_1(\Theta) + h_2(\Theta), $$
	where $h_1$ is a convex function, $h_2$ a differentiable and convex function and $L$ the gradient Lipschitz of $h_2$.
	
	\begin{algorithm}[H]
		\label{ProxMethod}
		\caption{Proximal gradient method} 
		\begin{algorithmic} 
			\STATE \textbf{Step $0$:} $\hat{\Theta}^{(0)}$ the null matrices. 
			\STATE\textbf{Step $t+1$:} $\hat{\Theta}^{(t+1)}=\textrm{prox}_{\lambda(1/L)h_1}(\hat{\Theta}^{(t)}-(1/L) \nabla h_2(\hat{\Theta}^{(t)}))$
		\end{algorithmic} 
	\end{algorithm}
	
	The main trick of the FISTA algorithm is to add a momentum term to the proximal gradient method, in order to yield smoother trajectory towards the convergence point. In addition, the proximal operator is performed on a specific linear combination of the previous two iterates, rather than on the previous iterate only. 
	
	\begin{algorithm}[H]
		\label{FISTA}
		\caption{FISTA (accelerated proximal gradient method)} 
		\begin{algorithmic} 
			\STATE \textbf{Step $0$:} $\kappa_0=0.1$, $\hat{\Theta}^{(0)}$ and $\Xi_0$ the null matrices. 
			\STATE \textbf{Step $t+1$:}
			\STATE \qquad \qquad \qquad $\hat{\Theta}^{(t+1)}=\textrm{prox}_{\lambda(1/L)h_1}(\Xi_t-(1/L) \nabla h_2(\Xi_t))$
			\STATE \qquad \qquad \qquad $\kappa_{k+1}=\frac{1+\sqrt{1+4\kappa_k^2}}{2}$
			\STATE \qquad \qquad \qquad $\Xi_{t+1}=\hat{\Theta}^{(t+1)}+\frac{\kappa_{k}-1}{\kappa_{k+1}}(\hat{\Theta}^{(t+1)}-\hat{\Theta}^{(t)})$
		\end{algorithmic} 
	\end{algorithm}
	
	In our specific model, to solve \eqref{OptimMAR}, $h_1(\Theta)=\|\Theta\|_\star$ and $h_2(\Theta)=\| \Omega \odot (\Theta-Y) \|^2_F$.
	Let us precise that: $$\frac{\partial h_2(\Theta)}{\partial \Theta_{ij}}=\nabla_{\Theta_{ij}} h_2(\Theta)=\Omega_{ij}\left(\Theta_{ij} - Y_{ij} \right).$$ Therefore, $$\nabla h_2(\Theta)=\Omega\odot (\Theta-Y)$$ and $L$ is equal to 1. 
	
	\section{\texttt{softImpute}}
	
	We start by describing \texttt{softImpute}. 
	\begin{algorithm}[H]
		\label{softImpute}
		\caption{\texttt{softImpute}}
		\begin{algorithmic} 
			\STATE \textbf{Step $0$:} $\hat{\Theta}^{(0)}$ the null matrix. 
			\STATE  \textbf{Step $t+1$:} $\hat{\Theta}^{(t+1)}=\textrm{prox}_{\lambda\|.\|_{\star}}(\Omega \odot Y + (1-\Omega) \odot \hat{\Theta}^{(t)})$
		\end{algorithmic} 
	\end{algorithm}
	
	\noindent
	The proximal operator of the nuclear norm of a matrix $X$ consists in a soft-thresholding of its singular values: we perform the SVD of $X$ and we obtain the matrices $U$, $V$ and $D$. Then $$\textrm{prox}_{\lambda\|.\|_{\star}}(X)=UD_{\lambda} V.$$ $D_{\lambda}$ is the diagonal matrix such that for all $i$, $$D_{\lambda,ii}=\max((\sigma_i-\lambda),0)$$, where the $(\sigma_{ii})$'s are the singular values of $X$.
	
	\subsection{Equivalence between \texttt{softImpute} and the proximal gradient method}\label{sec:Proxsoft}
	
	By using the same functions $h_1$ and $h_2$ as above, one has: 
	\begin{align*}
	\hat{\Theta}^{(t+1)}&=\textrm{prox}_{\lambda(1/L)h_1}(\hat{\Theta}^{(t)}-(1/L) \nabla h_2(\hat{\Theta}^{(t)})) \\
	&=\textrm{prox}_{\lambda\|.\|_\star}(\hat{\Theta}^{(t)}-\Omega \odot (\hat{\Theta}^{(t)} -Y)) \\
	&=\textrm{prox}_{\lambda\|.\|_\star}(\Omega \odot Y + (1-\Omega) \odot \hat{\Theta}^{(t)}),
	\end{align*}
	so that \texttt{softImpute} and the proximal gradient method are similar. 
	
	\subsection{Equivalence between the EM algorithm and iterative SVD in the MAR case}\label{sec:EMsoft}
	
	We prove here that in the MAR setting, \texttt{softImpute} is similar to the EM algorithm. Let us recall that in the MAR setting the model of the joint distribution is not needed but only the one of the data distribution, so that the E-step is written as follows: 
	
	\footnotesize
	\begin{align*}	Q(\Theta|\hat{\Theta}^{(t)})&=\mathbb{E}_{Y_\textrm{mis}}\left[\log(p(\Theta;y))|Y_\textrm{obs};\Theta=\hat{\Theta}^{(t)}\right] \\
	&\propto -\sum_{i=1}^{n} \sum_{j=1}^{p} \mathbb{E} \left[\left(\frac{y_{ij}-\Theta_{ij}}{\sigma}\right)^2 |\hat{\Theta_{ij}}^{(t)}\right],
	\end{align*}
	by using \eqref{eq:densitydata} and the independance of $Y_{ij}, \: \forall i,j$). Then, by splitting into the observed and the missing elements,
	\begin{multline*}
	Q(\Theta|\hat{\Theta}^{(t)})\propto-\sum_{i=1}^{n} \sum_{j, \: \Omega_{ij}=0} \mathbb{E}\left[ \left(\frac{y_{ij} -\Theta_{ij}}{\sigma}\right)^2 |\hat{\Theta_{ij}}^{(t)} \right]  \\
	-\sum_{i=1}^{n}\sum_{j, \: \Omega_{ij}=1} \left(\frac{y_{ij}-\Theta_{ij}}{\sigma}\right)^2 
	\end{multline*}
	Therefore,
	\begin{multline*}
	Q(\Theta|\hat{\Theta}^{(t)})\propto -\sum_{i=1}^{n} \sum_{j, \: \Omega_{ij}=0}\mathbb{E}[y_{ij} ^2 -2\Theta_{ij}y_{ij} +\Theta_{ij}^2| \hat{\Theta_{ij}}^{(t)} ]^2 \\
	-\sum_{i=1}^{n}\sum_{j, \: \Omega_{ij}=1} \left(\frac{y_{ij}-\Theta_{ij}}{\sigma}\right)^2
	\end{multline*}
	\begin{multline*}
	Q(\Theta|\hat{\Theta}^{(t)})\propto-\sum_{i=1}^{n} \sum_{j, \: \Omega_{ij}=0} \left(\sigma^2 + \hat{\Theta_{ij}}^{(t)2}-2\hat{\Theta_{ij}}^{(t)}\Theta_{ij}+\Theta_{ij}^2\right) \\ -\sum_{i=1}^{n}\sum_{j, \: \Omega_{ij}=1} \left(\frac{y_{ij}-\Theta_{ij}}{\sigma}\right)^2
	\end{multline*}
	which implies
	$Q(\Theta|\hat{\Theta}^{(t)})\propto \|\Omega \odot Y + (1-\Omega) \odot \hat{\Theta}^{(t)}-\Theta\|^2_F$
	
	\normalsize
	
	The M-step is then written as follows: 
	\[\hat{\Theta}^{(t+1)} \in \textrm{argmin}_{\Theta} \|\Omega \odot Y + (1-\Omega) \odot \hat{\Theta}^{(t)}-\Theta\|^2_F + \lambda\|\Theta\|_\star\]
	
	The proximal gradient method is applied with \[h_1(\Theta)=\lambda\|\Theta\|_\star \textrm{ and } h_2(\Theta)=\|\Omega \odot Y + (1-\Omega) \odot \hat{\Theta}^{(t)}-\Theta\|^2_F.\]
	Therefore, the EM algorithm in the MAR case is the same one as \texttt{softImpute}. 
	
	\section{The EM algorithm in the MNAR case}
	For the sake of clarity, we present below the EM algorithm in the MNAR and low dimension case.
	
	\begin{algorithm}[H]
		\label{EMMnarCase}
		\caption{The EM algorithm in the MNAR case} 
		\begin{algorithmic} 
			\STATE \textbf{Step $0$:} $\hat{\Theta}^{(0)}$ and $\hat{\phi}^{(0)}$.
			\STATE \textbf{Step $t+1$:}
			\FOR{$(i,j) \in \Omega_{\textrm{mis}}=\{ (l,k), l \in \left[1,n\right], k \in \left[1,p\right], \Omega_{lk}=0 \}$}
			\STATE{\textbf{draw} $z_{ij}^1, \dots, z_{ij}^{N_s} \overset{\textrm{i.i.d.}}{\sim} \left[y_{ij}|\Omega_{ij};\hat{\phi}^{(t)},\hat{\Theta}_{ij}^{(t)}\right]$ with the SIR algorithm. }
			\ENDFOR
			\STATE \textbf{Compute} $\hat{\Theta}^{(t+1)}$ by using \texttt{softImpute} or the FISTA algorithm with the imputed matrix $V$ (given by \eqref{matrixH}).
			\STATE \textbf{Compute} $\hat{\phi}^{(t+1)}$ by using the function glm with a binomial link, which perform a logistic regression of $J_{.1}^{(j)}$ on $J_{.2}^{(j)}$, with the matrix $J^{(j)}$ given above (\eqref{matrixJ}), for all $j$ such that $\exists i \in \{1,\dots n\}, \: \Omega_{ij}=0$.
		\end{algorithmic} 
	\end{algorithm}
	
	We already have given details for the stopping criterium. 
	
	We clarify the maximization step given by \eqref{max1} and \eqref{max2}. 
	
	\begin{align*}
	\hat{\Theta} &\in\underset{\Theta}{\textrm{argmin}} \sum_{i,j} \left( \frac{1}{N_s}\sum_{k=1}^{N_s} \frac{1}{2\sigma^2}(v_{ij}^{(k)}-\Theta_{ij})^2 \right) + \lambda\|\Theta\|_\star \\
	&\in \underset{\Theta}{\textrm{argmin}}\sum_{i,j}\left( \frac{1}{N_s\sigma^2}\sum_{k=1}^{N_s}-v_{ij}^{(k)}\Theta_{ij}+\frac{1}{2}\Theta_{ij}^2 \right) +\lambda\|\Theta\|_\star \\
	&\in \underset{\Theta}{\textrm{argmin}} \| V-\Theta \|_F^2 + \lambda \|\Theta\|_\star, \textrm{ where:}
	\end{align*} 
	\begin{equation}
	\label{matrixH}
	V=\begin{pmatrix}
	\frac{1}{N_s}\sum_{k=1}^{Ns}v_{11}^{(k)} \dots \frac{1}{N_s}\sum_{k=1}^{Ns}v_{1p}^{(k)} \\
	\vdots \ddots \vdots \\
	\frac{1}{N_s}\sum_{k=1}^{Ns}v_{n1}^{(k)} \dots \frac{1}{N_s}\sum_{k=1}^{Ns}v_{np}^{(k)}
	\end{pmatrix}
	\end{equation}
	
	\begin{align*}
	\hat{\phi}^{(t+1)} &\in \underset{\phi}{\textrm{argmin}} \sum_{i,j}\frac{1}{N_s} \sum_{k=1}^{N_s} (1-\Omega_{ij})C_3 - \Omega_{ij}C_4 \\
	&\in  \underset{\phi}{\textrm{argmin}}\sum_{i,j} \frac{1}{N_s} \sum_{k=1}^{N_s} C_3 + \Omega_{ij} \phi_{1j} (v_{ij}^{k}-\phi_{2j})
	\end{align*}
	with:
	\begin{align*}
	C_3&=\log(1+e^{-\phi_{1j}(v_{ij}^{k}-\phi_{2j})}) \\
	C_4&=\log(1-(1+e^{-\phi_{1j}(v_{ij}^{k}-\phi_{2j})})^{-1})
	\end{align*}
	
	For all $j \in \{1,\dots,p\}$ such that $\exists i \in \{1,\dots,n\}, \: \Omega_{ij}=0$, estimating the coefficients $\phi_{1j}$ and $\phi_{2j}$ remains to fit a generalized linear model with the binomial link function for the matrix $J^{(j)}$:  
	\begin{equation}
	\label{matrixJ}
	J^{(j)}=\begin{pmatrix}
	\Omega_{1j} & v_{1j}^{(1)}   \\
	\vdots & \vdots  \\
	\Omega_{nj} & v_{nj}^{(1)} \\ 
	\vdots & \vdots \\
	\Omega_{1j} & v_{1j}^{(N_s)}  \\
	\vdots & \vdots \\
	\Omega_{nj} & v_{nj}^{(N_s)}  \\
	\end{pmatrix}
	\end{equation}
	
	\subsection{SIR} \label{sec:SIR}
	In the Monte Carlo approximation, the distribution of interest is $\left[y_{ij}|\Omega_{ij};\hat{\phi}_j^{(t)},\hat{\Theta}_{ij}^{(t)}\right]$. By using the Bayes rules:
	\begin{align*}
	&p \left( y_{ij}|\Omega_{ij};\hat{\phi}_j^{(t)},\hat{\Theta}_{ij}^{(t)} \right) =: f(y_{ij}) \\
	&\propto p\left( y_{ij};\hat{\Theta}_{ij}^{(t)}\right)p\left(\Omega_{ij}|y_{ij};\hat{\phi}_j^{(t)}\right) =: g(y_{ij})
	\end{align*}
	Denoting the Gaussian density function of mean $\Theta^{(t)}_{ij}$ and variance $\sigma^2$ by $\varphi_{\Theta^{(t)}_{ij},\sigma^2}$, if $\sigma>(2\pi)^{-1/2}$, the following condition holds: \[f(y_{ij})=cg(y_{ij})\leq \varphi_{\Theta^{(t)}_{ij},\sigma^2}(x).\]
	
	For $M$ large, the SIR algorithm to simulate $$z \sim \left[y_{ij}|\Omega_{ij};\hat{\phi}_j^{(t)},\hat{\Theta}_{ij}^{(t)}\right]$$ is described as follows.
	\begin{algorithm}[H]
		\label{SIR}
		\caption{\texttt{SIR}}
		\begin{algorithmic} 
			\STATE \textbf{Draw:} a sample $x_1, \dots, x_M \sim \mathcal{N}(\Theta^{(t)}_{ij},\sigma^2)$.
			\STATE  \textbf{Compute the weights:} $$\omega(x_m)=\frac{g(x_m)}{\varphi_{\Theta^{(t)}_{ij},\sigma^2}(x_m)},$$  for  $m=1, \dots, M$.
			\STATE \textbf{Draw} $z$ from the original sample $x_1,\dots,x_M$ with probability proportional to $\omega(x_1),\dots,\omega(x_M)$.
		\end{algorithmic} 
	\end{algorithm}
	
	\section{Details on the variables in TraumaBase$^{\mbox{\normalsize{\textregistered}}}$}\label{sec:detailstrauma}
	
	A description of the variables which are used in Section \ref{sec:Traumabase} is given. The indications given in parentheses ph (pre-hospital) and h (hospital) mean that the measures have been taken before the arrival at the hospital and at the hospital.
	\begin{itemize}[label=\textbullet]
		\item \textit{SBP.ph}, \textit{DBP.ph}, \textit{HR.ph}: systolic and diastolic arterial pressure and heart rate during pre-hospital phase. (ph)
		\item \textit{HemoCue.init}: prehospital capillary hemoglobin concentration. (ph)
		\item \textit{SpO2.min}: peripheral oxygen saturation, measured by pulse oxymetry, to estimate oxygen content in the blood. (ph)
		\item \textit{Cristalloid.volume}: total amount of prehospital administered cristalloid fluid resuscitation (volume expansion). (ph)
		\item \textit{Shock.index.ph}: ratio of heart rate and systolic arterial pressure during pre-hospital phase. (ph)
		\item \textit{Delta.shock.index}: Difference of shock index between arrival at the hospital and arrival on the scene. (h)
		\item \textit{Delta.hemoCue}: Difference of hemoglobin level between arrival at the hospital and arrival on the scene. (h)
	\end{itemize}

\end{appendices}
	
	

	\newpage
	
	\section*{References}
	
	\bibliographystyle{plainnat}
	\bibliography{Article_LowRankMNARbib.bib}

\begin{thebibliography}{42}
\providecommand{\natexlab}[1]{#1}
\providecommand{\url}[1]{\texttt{#1}}
\expandafter\ifx\csname urlstyle\endcsname\relax
  \providecommand{\doi}[1]{doi: #1}\else
  \providecommand{\doi}{doi: \begingroup \urlstyle{rm}\Url}\fi

\bibitem[Audigier et~al.(2016)Audigier, Husson, and
  Josse]{audigier2016principal}
Vincent Audigier, Fran{\c{c}}ois Husson, and Julie Josse.
\newblock A principal component method to impute missing values for mixed data.
\newblock \emph{Advances in Data Analysis and Classification}, 10\penalty0
  (1):\penalty0 5--26, 2016.

\bibitem[Beck and Teboulle(2009)]{beck2009fast}
Amir Beck and Marc Teboulle.
\newblock A fast iterative shrinkage-thresholding algorithm for linear inverse
  problems.
\newblock \emph{SIAM journal on imaging sciences}, 2\penalty0 (1):\penalty0
  183--202, 2009.

\bibitem[Cai et~al.(2010)Cai, Cand{\`e}s, and Shen]{cai2010singular}
Jian-Feng Cai, Emmanuel~J Cand{\`e}s, and Zuowei Shen.
\newblock A singular value thresholding algorithm for matrix completion.
\newblock \emph{SIAM Journal on Optimization}, 20\penalty0 (4):\penalty0
  1956--1982, 2010.

\bibitem[Cai and Zhou(2013)]{cai2013max}
Tony Cai and Wen-Xin Zhou.
\newblock A max-norm constrained minimization approach to 1-bit matrix
  completion.
\newblock \emph{The Journal of Machine Learning Research}, 14\penalty0
  (1):\penalty0 3619--3647, 2013.

\bibitem[Candes and Plan(2010)]{candes2010matrix}
Emmanuel~J Candes and Yaniv Plan.
\newblock Matrix completion with noise.
\newblock \emph{Proceedings of the IEEE}, 98\penalty0 (6):\penalty0 925--936,
  2010.

\bibitem[Cand{\`e}s and Recht(2009)]{candes2009exact}
Emmanuel~J Cand{\`e}s and Benjamin Recht.
\newblock Exact matrix completion via convex optimization.
\newblock \emph{Foundations of Computational mathematics}, 9\penalty0
  (6):\penalty0 717, 2009.

\bibitem[Cand{\`e}s et~al.(2013)Cand{\`e}s, Sing-Long, and
  Trzasko]{candes2013sure}
Emmanuel~J Cand{\`e}s, Carlos~A Sing-Long, and Joshua~D Trzasko.
\newblock Unbiased risk estimates for singular value thresholding and spectral
  estimators.
\newblock \emph{IEEE Transactions on Signal Processing}, 61\penalty0
  (19):\penalty0 4643--4657, 2013.

\bibitem[Dempster et~al.(1977)Dempster, Laird, and Rubin]{dempster1977maximum}
Arthur~P Dempster, Nan~M Laird, and Donald~B Rubin.
\newblock Maximum likelihood from incomplete data via the em algorithm.
\newblock \emph{Journal of the Royal Statistical Society: Series B
  (Methodological)}, 39\penalty0 (1):\penalty0 1--22, 1977.

\bibitem[Gavish and Donoho(2017)]{gavish2017optimal}
Matan Gavish and David~L Donoho.
\newblock Optimal shrinkage of singular values.
\newblock \emph{IEEE Transactions on Information Theory}, 63\penalty0
  (4):\penalty0 2137--2152, 2017.

\bibitem[Gordon et~al.(1993)Gordon, Salmond, and Smith]{gordon1993novel}
Neil~J Gordon, David~J Salmond, and Adrian~FM Smith.
\newblock Novel approach to nonlinear/non-gaussian bayesian state estimation.
\newblock In \emph{IEE Proceedings F-radar and signal processing}, volume 140,
  pages 107--113. IET, 1993.

\bibitem[Harel and Schafer(2009)]{harel2009partial}
Ofer Harel and Joseph~L Schafer.
\newblock Partial and latent ignorability in missing-data problems.
\newblock \emph{Biometrika}, 96\penalty0 (1):\penalty0 37--50, 2009.

\bibitem[Hastie and Mazumder(2015)]{softManual}
Trevor Hastie and Rahul Mazumder.
\newblock \emph{softImpute: Matrix Completion via Iterative Soft-Thresholded
  SVD}, 2015.
\newblock URL \url{https://CRAN.R-project.org/package=softImpute}.
\newblock R package version 1.4.

\bibitem[Hastie et~al.(2015)Hastie, Mazumder, Lee, and Zadeh]{hastie2015matrix}
Trevor Hastie, Rahul Mazumder, Jason~D Lee, and Reza Zadeh.
\newblock Matrix completion and low-rank svd via fast alternating least
  squares.
\newblock \emph{The Journal of Machine Learning Research}, 16\penalty0
  (1):\penalty0 3367--3402, 2015.

\bibitem[Hay et~al.(2017)Hay, Abajobir, Abate, Abbafati, Abbas, Abd-Allah,
  Abdulkader, Abdulle, Abebo, Abera, et~al.]{hay2017global}
Simon~I Hay, Amanuel~Alemu Abajobir, Kalkidan~Hassen Abate, Cristiana Abbafati,
  Kaja~M Abbas, Foad Abd-Allah, Rizwan~Suliankatchi Abdulkader, Abdishakur~M
  Abdulle, Teshome~Abuka Abebo, Semaw~Ferede Abera, et~al.
\newblock Global, regional, and national disability-adjusted life-years (dalys)
  for 333 diseases and injuries and healthy life expectancy (hale) for 195
  countries and territories, 1990--2016: a systematic analysis for the global
  burden of disease study 2016.
\newblock \emph{The Lancet}, 390\penalty0 (10100):\penalty0 1260--1344, 2017.

\bibitem[Heckman(1974)]{heckman1974sample}
James~J Heckman.
\newblock Sample selection bias as a specification error.
\newblock \emph{Econometrica}, 42:\penalty0 679--94, 1974.

\bibitem[Ibrahim et~al.(1999)Ibrahim, Lipsitz, and Chen]{Ibrahim}
Joseph~G Ibrahim, Stuart~R Lipsitz, and M-H Chen.
\newblock Missing covariates in generalized linear models when the missing data
  mechanism is non-ignorable.
\newblock \emph{Journal of the Royal Statistical Society: Series B (Statistical
  Methodology)}, 61\penalty0 (1):\penalty0 173--190, 1999.

\bibitem[Josse and Husson(2012)]{josse2012selecting}
Julie Josse and Fran{\c{c}}ois Husson.
\newblock Selecting the number of components in principal component analysis
  using cross-validation approximations.
\newblock \emph{Computational Statistics \& Data Analysis}, 56\penalty0
  (6):\penalty0 1869--1879, 2012.

\bibitem[Josse et~al.(2016)Josse, Sardy, and Wager]{josse2016denoiser}
Julie Josse, Sylvain Sardy, and Stefan Wager.
\newblock denoiser: A package for low rank matrix estimation.
\newblock \emph{Journal of Statistical Software}, 2016.

\bibitem[Josse et~al.(2019)Josse, Prost, Scornet, and
  Varoquaux]{josse2019consistency}
Julie Josse, Nicolas Prost, Erwan Scornet, and Ga{\"e}l Varoquaux.
\newblock On the consistency of supervised learning with missing values.
\newblock \emph{arXiv preprint arXiv:1902.06931}, 2019.

\bibitem[Kallus et~al.(2018)Kallus, Mao, and Udell]{kallus2018causal}
Nathan Kallus, Xiaojie Mao, and Madeleine Udell.
\newblock Causal inference with noisy and missing covariates via matrix
  factorization.
\newblock \emph{arXiv preprint arXiv:1806.00811}, 2018.

\bibitem[Kishore~Kumar and Schneider(2017)]{kishore2017literature}
N~Kishore~Kumar and Jan Schneider.
\newblock Literature survey on low rank approximation of matrices.
\newblock \emph{Linear and Multilinear Algebra}, 65\penalty0 (11):\penalty0
  2212--2244, 2017.

\bibitem[Leek and Storey(2007)]{leek2007capturing}
Jeffrey~T Leek and John~D Storey.
\newblock Capturing heterogeneity in gene expression studies by surrogate
  variable analysis.
\newblock \emph{PLoS genetics}, 3\penalty0 (9):\penalty0 e161, 2007.

\bibitem[Little(1993)]{little1993pattern}
Roderick~JA Little.
\newblock Pattern-mixture models for multivariate incomplete data.
\newblock \emph{Journal of the American Statistical Association}, 88\penalty0
  (421):\penalty0 125--134, 1993.

\bibitem[Little and Rubin(2014)]{rubin_little}
Roderick~JA Little and Donald~B Rubin.
\newblock \emph{Statistical analysis with missing data}, volume 333.
\newblock John Wiley \& Sons, 2014.

\bibitem[Liu et~al.(2018)Liu, Dobriban, Singer, et~al.]{liu2018pca}
Lydia~T Liu, Edgar Dobriban, Amit Singer, et~al.
\newblock $ e $ pca: High dimensional exponential family pca.
\newblock \emph{The Annals of Applied Statistics}, 12\penalty0 (4):\penalty0
  2121--2150, 2018.

\bibitem[Mazumder et~al.(2010)Mazumder, Hastie, and
  Tibshirani]{mazumder2010spectral}
Rahul Mazumder, Trevor Hastie, and Robert Tibshirani.
\newblock Spectral regularization algorithms for learning large incomplete
  matrices.
\newblock \emph{Journal of machine learning research}, 11\penalty0
  (Aug):\penalty0 2287--2322, 2010.

\bibitem[Miao and Tchetgen~Tchetgen(2017)]{miao2017identification}
Wang Miao and E~Tchetgen~Tchetgen.
\newblock Identification and inference with nonignorable missing covariate
  data.
\newblock \emph{Statistica Sinica}, 2017.

\bibitem[Mohan and Pearl(2018)]{Mohan2018graphical}
Karthika Mohan and Judea Pearl.
\newblock Graphical models for processing missing data.
\newblock \emph{arXiv:1801.03583}, 2018.

\bibitem[Mohan et~al.(2018)Mohan, Thoemmes, and Pearl]{mohan2018estimation}
Karthika Mohan, Felix Thoemmes, and Judea Pearl.
\newblock Estimation with incomplete data: The linear case.
\newblock In \emph{IJCAI}, pages 5082--5088, 2018.

\bibitem[Morikawa et~al.(2017)Morikawa, Kim, and
  Kano]{morikawa2017semiparametric}
Kosuke Morikawa, Jae~Kwang Kim, and Yutaka Kano.
\newblock Semiparametric maximum likelihood estimation with data missing not at
  random.
\newblock \emph{Canadian Journal of Statistics}, 45\penalty0 (4):\penalty0
  393--409, 2017.

\bibitem[Murray(2018)]{murray}
Jared~S Murray.
\newblock Multiple imputation: A review of practical and theoretical findings.
\newblock \emph{arXiv preprint arXiv:1801.04058}, 2018.

\bibitem[Price et~al.(2006)Price, Patterson, Plenge, Weinblatt, Shadick, and
  Reich]{price2006principal}
Alkes~L Price, Nick~J Patterson, Robert~M Plenge, Michael~E Weinblatt, Nancy~A
  Shadick, and David Reich.
\newblock Principal components analysis corrects for stratification in
  genome-wide association studies.
\newblock \emph{Nature Genetics}, 38\penalty0 (8):\penalty0 904--909, 2006.

\bibitem[Robin et~al.(2018)Robin, Klopp, Josse, Moulines, and
  Tibshirani]{robin2018main}
Genevi{\`e}ve Robin, Olga Klopp, Julie Josse, {\'E}ric Moulines, and Robert
  Tibshirani.
\newblock Main effects and interactions in mixed and incomplete data frames.
\newblock \emph{arXiv preprint arXiv:1806.09734}, 2018.

\bibitem[Rubin(1976)]{rubin1}
Donald~B Rubin.
\newblock Inference and missing data.
\newblock \emph{Biometrika}, 63\penalty0 (3):\penalty0 581--592, 1976.

\bibitem[Rubin(2004)]{rubin2004multiple}
Donald~B Rubin.
\newblock \emph{Multiple imputation for nonresponse in surveys}, volume~81.
\newblock John Wiley \& Sons, 2004.

\bibitem[Seaman et~al.(2013)Seaman, Galati, Jackson, and Carlin]{Seaman2013MAR}
Shaun Seaman, John Galati, Dan Jackson, and John Carlin.
\newblock What is meant by missing at random?
\newblock \emph{Statist. Sci.}, 28\penalty0 (2):\penalty0 257--268, 05 2013.

\bibitem[Tang and Ishwaran(2017)]{tang2017random}
Fei Tang and Hemant Ishwaran.
\newblock Random forest missing data algorithms.
\newblock \emph{Statistical Analysis and Data Mining: The ASA Data Science
  Journal}, 10\penalty0 (6):\penalty0 363--377, 2017.

\bibitem[Twala et~al.(2008)Twala, Jones, and Hand]{twala2008good}
BETH Twala, MC~Jones, and David~J Hand.
\newblock Good methods for coping with missing data in decision trees.
\newblock \emph{Pattern Recognition Letters}, 29\penalty0 (7):\penalty0
  950--956, 2008.

\bibitem[Udell and Townsend(2017)]{Udell2017logrank}
Madeleine Udell and Alex Townsend.
\newblock Nice latent variable models have log-rank.
\newblock \emph{ArXiv}, abs/1705.07474, 2017.
\newblock URL \url{http://arxiv.org/abs/1705.07474}.

\bibitem[Udell et~al.(2016)Udell, Horn, Zadeh, Boyd,
  et~al.]{udell2016generalized}
Madeleine Udell, Corinne Horn, Reza Zadeh, Stephen Boyd, et~al.
\newblock Generalized low rank models.
\newblock \emph{Foundations and Trends{\textregistered} in Machine Learning},
  9\penalty0 (1):\penalty0 1--118, 2016.

\bibitem[Verbanck et~al.(2015)Verbanck, Josse, and
  Husson]{verbanck2015regularised}
Marie Verbanck, Julie Josse, and Fran{\c{c}}ois Husson.
\newblock Regularised pca to denoise and visualise data.
\newblock \emph{Statistics and Computing}, 25\penalty0 (2):\penalty0 471--486,
  2015.

\bibitem[Yang et~al.(2018)Yang, Akimoto, Kim, and Udell]{yang2018oboe}
Chengrun Yang, Yuji Akimoto, Dae~Won Kim, and Madeleine Udell.
\newblock Oboe: Collaborative filtering for automl initialization.
\newblock \emph{arXiv preprint arXiv:1808.03233}, 2018.

\end{thebibliography}
	
	
	
	
	
	

\end{document}